\documentclass[a4paper,fleqn]{cas-dc}

\usepackage[authoryear,longnamesfirst]{natbib}
\usepackage{graphicx}
\usepackage{amsmath}
\usepackage{amssymb}
\usepackage{booktabs}
\usepackage{arydshln}
\usepackage{multirow}
\usepackage{float}
\usepackage{orcidlink}
\usepackage[capitalize]{cleveref}
\crefname{section}{Sec.}{Secs.}
\Crefname{section}{Section}{Sections}
\Crefname{table}{Table}{Tables}
\crefname{table}{Tab.}{Tabs.}

\usepackage[inline]{enumitem}
\usepackage{amsthm}
\usepackage{amsmath}
\usepackage{graphicx}
\usepackage[table]{xcolor}
\usepackage{bbm}
\usepackage{amssymb}
\usepackage{xcolor}
\usepackage{caption}
\usepackage{xspace}
\usepackage{pifont}
\usepackage{booktabs}
\usepackage{bm}
\usepackage{subcaption} 
\newcommand{\methodname}{\texttt{TgRA}\xspace}
\newcommand{\othermethod}{\texttt{TRADES-U}\xspace}

\begin{document}
\let\WriteBookmarks\relax
\def\floatpagepagefraction{1}
\def\textpagefraction{.001}

\shorttitle{Learning Robustness at Test-Time from a Non-Robust Teacher}
\shortauthors{Bianchettin et al.}

\title[mode=title]{Learning Robustness at Test-Time from a Non-Robust Teacher}



\author[1]{Stefano Bianchettin\orcidlink{0009-0001-1783-0727}}
\ead{s.bianchettin@santannapisa.it}

\author[1]{Giulio Rossolini\orcidlink{0000-0002-6404-2627}}
\ead{g.rossolini@sssup.it}

\author[1]{Giorgio Buttazzo\orcidlink{0000-0003-4959-4017}}
\ead{g.buttazzo@sssup.it}

\affiliation[1]{organization={Department of Excellence in Robotics and AI, Sant'Anna School of Advanced Study},
             addressline={Via Moruzzi, 1},
             city={Pisa},
             postcode={56124},
             state={Tuscany},
             country={Italy}}

\begin{abstract}
Nowadays, pretrained models are increasingly used as general-purpose backbones and adapted at test-time to downstream environments where target data are scarce and unlabeled. While this paradigm has proven effective for improving clean accuracy on the target domain, adversarial robustness has received far less attention, especially when the original pretrained model is not explicitly designed to be robust. This raises a practical question: \emph{can a pretrained, non-robust model be adapted at test-time to improve adversarial robustness on a target distribution?}
To face this question, this work studies how adversarial training strategies behave when integrated into adaptation schemes for the unsupervised test-time setting, where only a small set of unlabeled target samples is available. It first analyzes how classical adversarial training formulations can be extended to this scenario, showing that straightforward distillation-based adaptations remain unstable and highly sensitive to hyperparameter tuning, particularly when the teacher itself is non-robust. 

To address these limitations, the work proposes a label-free framework that uses the predictions of a non-robust teacher model as a semantic anchor for both the clean and adversarial objectives during adaptation. We further provide theoretical insights showing that our formulation yields a more stable alternative to the self-consistency-based regularization commonly used in classical adversarial training.
Experiments evaluate the proposed approach on CIFAR-10 and ImageNet under induced photometric transformations. The results support the theoretical insights by showing that the proposed approach achieves improved optimization stability, lower sensitivity to parameter choices, and a better robustness-accuracy trade-off than existing baselines in this post-deployment test-time setting.

\end{abstract}


\maketitle

 \section{Introduction}
\label{sec:intro}
Deep neural networks have reached remarkable levels of performance and generalization, enabling their deployment in a wide range of real-world applications where operating conditions are only partially known at design time \cite{zeng2024rethinking,adapt_NGUYENMEIDINE2021104096, adapt_SHAMSOLMOALI2021104268}. In practice, large-scale pretrained models are increasingly adopted as generic standalone backbones, which are then fine-tuned or adapted to downstream domains where only small datasets are available. This paradigm allows deploying pretrained models in new environments, even when data are limited, difficult to collect, and unlabeled.

However, current generalization and adaptation practices mainly rely on clean accuracy, while do not necessarily translate into adversarial robustness \cite{goodfellow2014explaining,madry2017towards}, especially when pretrained models are not explicitly designed to be robust. Even mild distribution shifts, arising from changes in sensing conditions, data acquisition pipelines, or environmental factors, can significantly degrade a model’s robustness to adversarial inputs~\cite{hendrycks2019benchmarking, taori2020measuring}, which provides an important alert for the use both in security and safety critical applications, where vulnerabilities to adversarial perturbations may have severe consequences~\cite{amodei2016concrete}.

These practical trends raise a fundamental and concrete question that places robustness under the lens of test-time adaptation: given a pretrained, non-robust model deployed in a new environment, 
\textit{how its robustness on the target distribution can be improved, without having access to the original large-scale source data and relying only on a limited set of unlabeled target samples.}
Addressing this problem is non-trivial, since standard adversarial training methods, such as PGD~\cite{madry2017towards} or TRADES~\cite{zhang2019theoretically}, are not directly applicable, as they require ground-truth labels. Similarly, many distillation-based alternatives rely on predictions from a robust teacher, which may not be available in many real-world applications. These limitations motivate the need for robust adaptation methods tailored specifically to unlabeled target environments.

\begin{figure*}[t]
    \centering
    \includegraphics[width=0.95\textwidth]{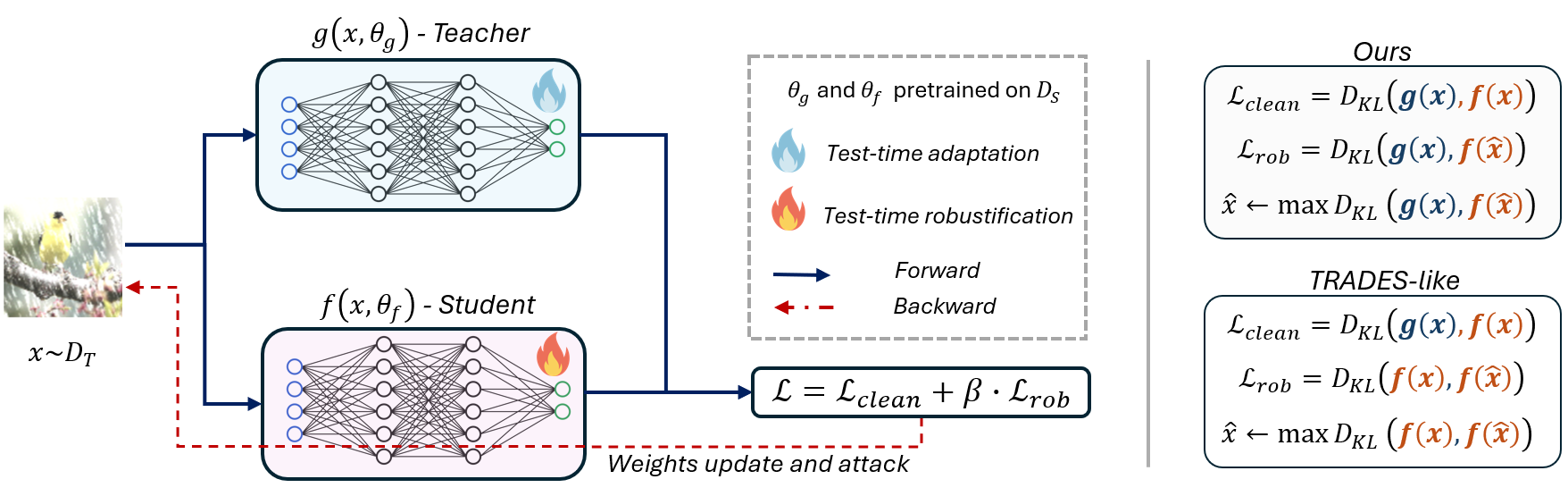}
    \caption{
    Overview of the proposed Teacher-guided Robust Adaptation (\methodname{}) framework, detailed in Section~\ref{sec:method}: the student model is fine-tuned at test-time under distribution shift, using a non-robust teacher as a reference for both the accuracy and robustness objectives, unlike TRADES-like approaches.}
    \label{fig:intro}
\end{figure*}

\medskip
\noindent \emph{This work}. 
The goal of this work is to investigate whether a pretrained model can be adapted after deployment to improve adversarial robustness on a target distribution, under a realistic setting where neither labeled target data, nor source data, nor adversarially robust pretrained models are available.
As a first step, we study how classical adversarial training schemes can be extended to this setting. A natural baseline is to replace the supervised accuracy term in adversarial training with a distillation objective driven by a teacher model that retains good clean generalization on the target domain~\cite{chen2021ltd}. However, this formulation, while feasible, remains prone to training instability and is highly sensitive to hyperparameter tuning. 

To address this limitation, this work proposes \emph{Teacher-guided Robust Adaptation} (\methodname), a framework that extends adversarial training by using soft targets from a non-robust teacher model to guide \emph{both} the clean and the robust objectives (illustration and description in Figure~\ref{fig:intro}). This formulation is further supported by theoretical insights, showing that anchoring both terms to a teacher provides a more stable and effective alternative to self-consistency-based formulations.

The experimental analysis evaluates the proposed approach on CIFAR-10 and ImageNet-Val under simulated target scenarios, considering different architectures and adaptation settings. Compared with existing baselines and alternative strategies, \methodname{} shows improved training stability, a reduced parameter sensitivity and a better robustness-accuracy trade-off on the target distribution. More broadly, we hope this work encourages further investigation into practical adversarial training strategies for realistic post-deployment adaptation scenarios.
In summary, this work provides the following contributions:
\begin{itemize}
    \item It extends the classic paradigm of adversarial training to face the problem of adversarial robustness adaptation in unsupervised test-time settings, where a non-robust pretrained model must be adapted to improve robustness under the target domain;
    \item It proposes and theoretically supports \methodname{}, a label-free test-time robustness adaptation framework, in which a non-robust teacher is used as a semantic anchor for both the clean and adversarial objectives;
    \item Through extensive experiments, it provides empirical insights into this setting and shows that \methodname{} consistently improves the robustness-accuracy trade-off across the addressed scenarios.
\end{itemize}

\textbf{Paper structure.}
The remainder of the paper is organized as follows. Section~\ref{sec:related} reviews state-of-the-art methods and highlights the motivations behind the work. Section~\ref{sec:background} introduces the necessary background and formalizes the problem. Section~\ref{sec:adversarial_training_robust} examines classical adversarial training formulations in the test-time adaptation setting. Section~\ref{sec:method} introduces \methodname{} and presents theoretical insights. Section~\ref{sec:experiments} reports the experimental results. Finally, Section~\ref{sec:conclusions} concludes the paper and outlines directions for future research. \emph{The code is publicly available}\footnote{\url{https://github.com/stefanobianco12/learning_robustness_test_time}}.

 \section{Related work}
\label{sec:related}
\paragraph{Adversarial training and robust fine-tuning.}
Adversarial training is the standard approach for achieving robustness against norm-bounded perturbations, where the main objective is to solve a training-time min--max optimization problem between model robustification and attack effectiveness~\cite{madry2017towards}. This idea was further improved by TRADES~\cite{zhang2019theoretically}, which explicitly addresses the trade-off between clean accuracy and robustness through a KL-based regularizer. Several variants of this approach were later proposed to refine the supervised TRADES formulation, for example by revisiting the role of misclassified samples~\cite{wang2019improving} or the form of the KL objective itself~\cite{cui2024decoupled}. While most works address the problem of training a robust model from scratch, other studies have explored robust fine-tuning on the same training dataset, introducing an adversarial training scheme after some epochs to reduce computational cost ~\cite{jeddi2020simple,zhu2023improving,ngnawe2025robust}.

In addition to the above techniques, several works have explored the use of knowledge distillation~\cite{hinton2015distilling,furlanello2018born} in adversarial training. In these settings, it is typically assumed that a robust teacher model is already available and can provide distillation-based supervision to a student model, with the goal of transferring robustness from teacher to student~\cite{dong2024adversarially,zhu2021reliable}. However, these works assume the availability of a teacher that is robust a priori, which may limit the practical applicability of the scheme in test-time robustification settings. To the best of our knowledge, only Chen et al.~\cite{chen2021ltd} propose a training approach based on a non-robust teacher, where it is used  to provide soft labels to the student for clean-accuracy supervision in a training-time setting, while the student still relies on a self-consistency 
robustness term, as in the TRADES formulation.
Importantly, all of the methods discussed above still rely on fully labeled source data, which limits their applicability to test-time scenarios.

\paragraph{Test-time Adaptation under distribution shift.}
A central practical challenge for pretrained task models is maintaining reliability when the test-time distribution differs from the one seen during training, often through changes in appearance or other domain characteristics~\cite{hendrycks2019benchmarking, pietrosanti2025benchmarking}. In this setting, an important problem studied in the literature is how to adapt a deployed model when source data are no longer available, and only samples from the target domain can be accessed. Test-time adaptation (TTA) addresses this challenge, where early work introduced test-time training as a self-supervised update performed on each test sample or mini-batch~\cite{sun2020test, wang2020tent, zhang2022memo, wang2022continual}. Overall, this literature is closely aligned with our deployment assumptions, but not with our objective: these methods are primarily designed to recover clean accuracy under shift without enabling adversarial robustness within the addressed target domain, which, from a security perspective, is crucial.

More closely related to adversarial robustness, Alhamoud et al.~\cite{alhamoud2022generalizability, pietrosanti2025benchmarking} show that adversarial robustness can degrade substantially on unseen domains, and that this degradation is not well predicted by simple input-level similarity measures. Nevertheless, only a smaller body of work has examined TTA explicitly from a robustness perspective. Croce et al.~\cite{croce2022evaluating} evaluate adaptive test-time defenses under carefully designed attacks and show that many such defenses do not outperform static alternatives under stricter evaluation. Recent studies have also examined the attack surface introduced by adaptation itself, either by revealing poisoning vulnerabilities in TTA pipelines or by modifying batch-statistics estimation to reduce sensitivity to malicious samples~\cite{lo2024adaptive,park2024medbn}. These works are particularly relevant here because they highlight both the potential and the fragility of inference-time adaptation, and they motivate the need for test-time learning strategies that remain robust under adversarial conditions.

\paragraph{Positioning of our work.}
The setting addressed in this paper lies at the intersection of these lines of research to understand how robustification, particularly through adversarial training, can support practical test-time scenarios, where models are typically deployed as pretrained, non-robust networks and retraining from scratch is often infeasible. This makes fine-tuning a natural and practically important setting in which to study robustness ~\cite{jeddi2020simple,zhu2023improving,ngnawe2025robust}, especially when only limited target data is available ~\cite{ngnawe2025robust}.
Furthermore, the need to address this setting is also reinforced by the fact that robustness is highly sensitive to distribution shift~\cite{alhamoud2022generalizability} 
when even mild test-time shifts applied to common pretrained models substantially degrade adversarial robustness, while leaving clean performance relatively stable.

\section{Background and formalism}
\paragraph{Background.}
\label{sec:background}
Let $\mathcal{D}$ denote a data distribution over input–label pairs $(x,y)$, where $x \in \mathcal{X}$ and $y \in \mathcal{Y}$, with $\mathcal{X}$ the input space and $\mathcal{Y}$ the output class space. 
Let $f(\cdot;\theta_f): \mathcal{X} \rightarrow \mathcal{Y}$ be a neural network parameterized by $\theta_f$, trained on a source dataset $\mathcal{D}_S$.
At deployment time, the model is evaluated on samples drawn from a different distribution $\mathcal{D}_T \neq \mathcal{D}_S$, while the former is not more available.

Given a norm $p$, the $p$-ball of radius $\varepsilon > 0$ centered at $x$ is
\(
\mathcal{B}_{\varepsilon,p}(x) = \left\{ \hat{x} \in \mathcal{X} \;:\; \|\hat{x} - x\|_p \le \varepsilon \right\},
\)
and an adversarial example is an input $\hat{x} \in \mathcal{B}_{\varepsilon,p}(x)$ such that $f(\hat{x};\theta) \neq y$, while the original input $x$ is correctly classified, i.e., $f(x;\theta) = y$.
We define the clean error $\mathcal{R}_{\mathrm{clean}}(f, \mathcal{D})$ and the adversarial error $\mathcal{R}_{\mathrm{rob}}(f, \mathcal{D}, \varepsilon)$ on a generic distribution $\mathcal{D}$ as
\[
\mathcal{R}_{\mathrm{clean}}(f, \mathcal{D}) =
\mathbb{E}_{(x,y)\sim\mathcal{D}} \left[ \mathbb{I}\big(f(x;\theta_f) \neq y\big) \right],
\]
\[
\mathcal{R}_{\mathrm{rob}}(f, \mathcal{D}, \varepsilon) =
\mathbb{E}_{(x,y)\sim\mathcal{D}} \left[
\max_{\hat{x} \in \mathcal{B}_{\varepsilon,p}(x)}
\mathbb{I}\big(f(\hat{x};\theta_f) \neq y\big)
\right].
\]

\paragraph{Problem Setting.}
We consider a model $f(\cdot;\theta_f)$ trained on a source distribution $\mathcal{D}_S$, and deployed on a target distribution $\mathcal{D}_T \neq \mathcal{D}_S$. At deployment time, only unlabeled samples $\{x\} \sim \mathcal{D}_T$ are accessible, while the source data $\mathcal{D}_S$ is no longer available.
The objective is to adapt the model parameters $\theta_f$ at test time to both improve adversarial robustness and clean accuracy on $\mathcal{D}_T$:
\begin{equation}
\label{eq:goal_test_time}
\min_{\theta_f} \; 
\mathcal{R}_{\text{clean}}(f, \mathcal{D}_T)
+ 
\mathcal{R}_{\text{rob}}(f, \mathcal{D}_T, \varepsilon).
\end{equation}
Note that this objective cannot be addressed using labels 
$y$, which are unavailable at test time; consequently, it is necessary to consider a fine-tuning setting that differs from the standard adversarial training frameworks commonly studied in the literature.

\section{Adversarial Training in Robust Adaptation}
\label{sec:adversarial_training_robust}
Before introducing the proposed approach, this section analyzes how standard adversarial training strategies behave when adapted to the unsupervised test-time setting. 

\paragraph{Classic Adversarial Training.} 
Adversarial training can be generally interpreted as optimizing a trade-off between clean accuracy and robustness in a supervised setting. As for Equation \eqref{eq:goal_test_time}, this objective is commonly formulated as a bi-level optimization problem, in which an inner maximization constructs adversarial perturbations within $\mathcal{B}_{\varepsilon,p}(x)$, and an outer minimization updates the model parameters:
\begin{equation}
\min_{\theta_f} \;
\mathbb{E}_{(x,y)\sim \mathcal{D}}
\Big[
\underbrace{\mathcal{L}_\text{acc}(f(x;\theta_f), y)}_{\text{accuracy term}}
+ 
\beta \cdot
\underbrace{
\max_{\hat{x} \in \mathcal{B}_{\varepsilon,p}(x)}
\mathcal{L}_\text{rob}(f(\hat{x};\theta_f), y)
}_{\text{robustness term}}
\Big].
\end{equation}
\noindent Here, $\mathcal{D}$ denotes a generic distribution used for training. Several methods instantiate these two terms differently. For instance, standard PGD-based adversarial training \cite{madry2017towards} uses cross-entropy for both clean and adversarial inputs, i.e., $\mathcal{L}_{\mathrm{rob}} = \ell_{\mathrm{CE}}(f(\hat{x};\theta), y)$, while TRADES~\cite{zhang2019theoretically} decouples the two objectives by combining a supervised accuracy term with a self-consistency-based robustness regularization:
\begin{equation}
\label{eq:trades}
\mathcal{L}_{\mathrm{acc}} = \ell_{\mathrm{CE}}(f(x;\theta), y), \qquad
\mathcal{L}_{\mathrm{rob}} = D_{\mathrm{KL}}\big(f(x;\theta)\,\|\,f(\hat{x};\theta)\big).
\end{equation}
where $D_{\mathrm{KL}}$ is the Kullback-Leibler logit divergence. This formulation has been shown to improve the accuracy-robustness trade-off and has inspired other techniques ~\cite{wang2019improving,cui2024decoupled}.

\paragraph{Adversarial Training in Test-time Robust Adaptation.}
Adapting the above frameworks to unsupervised test-time settings is not straightforward. The primary challenge is that the accuracy term inherently depends on ground-truth labels, which provide a critical anchor for the optimization process. In the absence of labels $y$, the robustness term alone is generally insufficient to guide learning and may lead to unstable weight updates \cite{zhang2019theoretically}.
To address these issues, a natural strategy is to replace the supervised accuracy term with a knowledge distillation objective (e.g., LTD~\cite{chen2021ltd}, STARSHIP~\cite{dong2024adversarially}), where a teacher model $g(\cdot;\theta_g)$, expected to well generalize with the target distribution, provides soft targets to guide a student model $f(\cdot;\theta_f)$.

Following this direction, this paper considers an unsupervised variant of TRADES (denoted as \othermethod) specifically applied on the target domain $\mathcal{D}_T$ for test-time settings: 
\begin{equation}
\label{eq:trades-u}
\begin{aligned}
\min_{\theta_f} \;
\mathbb{E}_{x \sim \mathcal{D}_T} \Big[
&D_{\mathrm{KL}}\!\left(
g(x;\theta_g) \;\|\; f(x;\theta_f)
\right) \\
&+ \beta \cdot
\max_{\hat{x} \in \mathcal{B}_{\varepsilon,p}(x)}
D_{\mathrm{KL}}\!\left(
f(x;\theta_f) \;\|\; f(\hat{x};\theta_f)
\right)
\Big].
\end{aligned}
\end{equation}

Note that the teacher $g(\cdot;\theta_g)$, although originally trained on the source distribution as the student $f(\cdot;\theta_f)$, is not treated as a fully frozen model. Instead, it is adapted before or in parallel with the target distribution to improve its clean (non-robust) accuracy. For this purpose, a standard test-time adaptation strategy can be applied to the teacher, while robustness adaptation is carried out on the student model.

While the above formulation is fully label-free, it also introduces a key limitation. Specifically, the robustness term still depends on the self-consistency of the student model \(f(\cdot;\theta_f)\), where, in the fine-tuning scenario considered in this work, this may result in unstable optimization dynamics and increased sensitivity to the parameter \(\beta\), as theoretically discussed in Section~\ref{sec:method}.

\section{Methodology}
\label{sec:method}
To provide a better formulation of adversarial training, this section proposes \methodname, a simple yet effective variant of \othermethod that reduces divergence and instability during unsupervised robustness adaptation on $\mathcal{D}_T$. The key  idea, illustrated in Figure \ref{fig:intro}, is to explicitly leverage the non-robust teacher model as a reference not only for the accuracy term, but also for the robustness term:
\begin{equation}
\label{eq:kl_reference}
\begin{aligned}
\min_{\theta_f} \; \mathbb{E}_{x \sim \mathcal{D}_T} \Big[ \;&
D_{\mathrm{KL}}\!\left(
g(x;\theta_g) \;\|\; f(x;\theta_f)
\right) \\
&+ \beta \cdot
\max_{\hat{x} \in \mathcal{B}_{\varepsilon,p}(x)}
D_{\mathrm{KL}}\!\left(
g(x;\theta_g) \;\|\; f(\hat{x};\theta_f)
\right)
\Big].
\end{aligned}
\end{equation}
Here, adversarial examples $\hat{x}$ are generated by explicitly maximizing the divergence between the student’s predictions and the teacher’s predictions on clean inputs.
The first term, the same term used in Equation~\eqref{eq:trades-u} for \othermethod, aligns the student and teacher predictions on clean samples, while the second term, which enforces robustness by requiring the student to remain consistent with the teacher’s clean predictions even under worst-case perturbations. 

Importantly, as will be formalized below, the key benefit of this design is that the teacher model is not subject to adversarial optimization. Its outputs, therefore, remain consistent, making it significantly less susceptible to change and instability during optimization. As a result, the proposed formulation yields a more stable optimization process for the student and reduces sensitivity to the hyperparameter~$\beta$.

\subsection{Theoretical analysis}
\label{sec:theory_stability}
This section provides a formal perspective on why anchoring both the clean and adversarial objectives to the teacher's output distribution can stabilize optimization under domain shift, in contrast to the student self-consistency regularization used in TRADES. 
Note that, in the analysis below, we treat the teacher output distribution as fixed during the student update. However, the same formulation can also be interpreted as a local approximation when the teacher is adapted in parallel.
Let $p_{\theta}(x)$ denote the predictive distribution of the student, i.e., $\text{softmax}(f(x))$, $\theta$ its weights, and let $q(x)$ denote the predictive distribution of the teacher, i.e., $\text{softmax}(g(x))$. Consider the following two regularizers:
\begin{align}
R_{\mathrm{self}}(\theta;x)
&=
\max_{\hat{x}\in\mathcal{B}_{\varepsilon,p}(x)}
\mathrm{KL}\!\big(p_\theta(x)\,\|\,p_\theta(\hat{x})\big), \\
R_{\mathrm{teach}}(\theta;x)
&=
\max_{\hat{x}\in\mathcal{B}_{\varepsilon,p}(x)}
\mathrm{KL}\!\big(q(x)\,\|\,p_\theta(\hat{x})\big).
\end{align}
\noindent Here, $\mathrm{KL}$ denotes the standard Kullback--Leibler divergence between output probability distributions. These two regularizers correspond to the robustness terms adopted in \othermethod \eqref{eq:trades-u} and \methodname \eqref{eq:kl_reference}, respectively, and can be analyzed mathematically to highlight their different optimization dynamics.
The two regularizers admit the following gradients.

\paragraph{Proposition.} The gradient of the self-consistency regularizer admits the following decomposition:
\begin{equation}
\label{eq:self_grad_main}
\begin{aligned}
\nabla_\theta R_{\mathrm{self}}(\theta;x)
={}&
\underbrace{
\big(\partial_\theta p_\theta(x)\big)^\top
\nabla_{p} \mathrm{KL}\!\big(p\,\|\,p_\theta(\hat{x}_\theta)\big)\big|_{p=p_\theta(x)}
}_{\text{reference-side term}}
\\[2pt]
&+
\underbrace{
\big(\partial_\theta p_\theta(\hat{x}_\theta)\big)^\top
\nabla_{p'} \mathrm{KL}\!\big(p_\theta(x)\,\|\,p'\big)\big|_{p'=p_\theta(\hat{x}_\theta)}
}_{\text{adversarial-side term}} .
\end{aligned}
\end{equation}
whereas the teacher-anchored regularizer satisfies:
\begin{equation}
\nabla_\theta R_{\mathrm{teach}}(\theta;x)
=
\big(\partial_\theta p_\theta(\hat{x}_\theta)\big)^\top
\nabla_{p'} \mathrm{KL}\!\big(q(x)\,\|\,p'\big)\big|_{p'=p_\theta(\hat{x}_\theta)}.
\label{eq:teach_grad_main}
\end{equation}
A full derivation and additional analysis are provided in Appendix~\ref{sec:derivation_prop}. These expressions form the basis for the stability analysis discussed in the following.

\paragraph{ \noindent\textbf{•} Stability of the Optimization Dynamics:}
Considering the above derivation, $R_{\mathrm{self}}$ depends on $\theta$ through both the clean reference and the adversarial prediction, while $R_{\mathrm{teach}}$ depends on $\theta$ only through the adversarial branch.
This shows that self-consistency introduces a kind of {moving-target} effect: the reference distribution $p_\theta(x)$ changes during optimization, and its variation (not negligible, as discussed next) directly affects the gradient through the reference-side term in Equation~\eqref{eq:self_grad_main}. In contrast, teacher anchoring removes this coupling by replacing the moving reference with a consistent target $q(x)$, as shown in Equation~\eqref{eq:teach_grad_main}.

\paragraph{\noindent\textbf{•} Stability with Respect to $\beta$:}
In the original TRADES formulation, the parameter $\beta$ was designed to control the trade-off between clean accuracy and robustness, i.e. $\mathcal{L}(\theta)=\mathcal{L}_{\mathrm{acc}}(\theta)+\beta\,R_\text{self}(\theta;x)$. However,  $\beta$ could also affect the stability of the optimization dynamics. Specifically, in the self-consistency case, increasing $\beta$ amplifies both the adversarial-side term and the reference-side term in Equation \eqref{eq:self_grad_main}. Since the reference distribution itself depends on the student parameters, this scaling may magnify optimization instability. By contrast, in the teacher-anchored case, $\beta$ scales only the adversarial alignment to a more stable target. As a result, although $\beta$ remains effective in controlling the trade-off, the stability of the proposed formulation is expected to be less sensitive to its choice.

\paragraph{\noindent\textbf{•} Stability in Robust Fine-Tuning:}
Note that the instability induced by self-consistency is particularly critical under non-robust initialization, which is the setting considered in this test-time analysis. At the beginning of adaptation, the student is initialized from the pretrained model, so that $p_\theta(x) \approx q(x)$. In this regime, the robustness loss may have a stronger impact than the accuracy term.
Therefore, the self-consistency regularizer strongly enforces invariance of $p_\theta(x)$ under adversarial perturbations. However, since the model is not robust at initialization, the adversarial examples $\hat{x}$ can induce large and misaligned gradients that dominate the update and rapidly distort the clean predictions $p_\theta(x)$.
As a consequence, the reference distribution in the self-consistency term begins to drift early during optimization. This effect can lead to instability and significant performance degradation in the initial stages of adaptation, as shown in the experimental analysis in Section~\ref{sec:experiments}. In contrast, teacher anchoring maintains a consistent reference $q(x)$ throughout optimization. This prevents the robustness term from altering the clean prediction target and avoids the collapse induced by early adversarial updates, resulting in more stable adaptation dynamics.

\section{Experimental results}
\label{sec:experiments}
This section presents and discusses the experimental results, structured around the following research questions: 
\emph{(i)} Does the proposed \methodname{} yield greater improvements and more stable behavior than the unsupervised \othermethod{} baseline? \emph{(ii)} How does the method behave as the amount of unlabeled target data varies? and \emph{(iii)} To what extent can a label-free, teacher-anchored strategy approach the performance of supervised adversarial fine-tuning methods?
Across these settings, the experiments report both clean and robust accuracy under adversarial evaluation, with the goal of assessing the robustness-accuracy trade-off as the target-domain shift increases.

\subsection{Experimental setup}
\label{sec:exp_setings}
\noindent \textbf{Datasets.}
The proposed method is evaluated on CIFAR-10 and ImageNet under distribution shift induced by corrupted target samples. For CIFAR-10, we follow the approach in~\cite{hendrycks2019benchmarking}, considering multiple corruption types, including Gaussian noise, Gaussian blur, and color jitter, with severity levels from $0$ (clean) to $2$. The exact corruption pipeline is reported in Appendix~\ref{app:corruptions}. We primarily consider target domains derived from the CIFAR-10 training set. To analyze different adaptation distributions, we also examine a subset of the CIFAR-10 test set in Sections~\ref{sec:target_samples} and~\ref{sec:supervised}. In the first case, the evaluation part is performed on the corrupted CIFAR-10 test set; in the second, on the remaining disjoint test samples.
For ImageNet, we use the validation set as target-domain data and split it into two balanced subsets, one for adaptation and one for evaluation. Random seeds and additional technical details are provided in the public repository (see Section~\ref{sec:intro}).

\medskip
\noindent \textbf{Models.} 
For CIFAR-10, both the teacher and the student are initialized from the same pretrained WideResNet-34~\cite{zagoruyko2016wide}, which is trained on the original CIFAR-10 training set.
For ImageNet, we consider two pretrained architectures: ResNet-50 \cite{he2016deep}, and ViT-B/16 \cite{dosovitskiy2020image}.
Details about the pretrained weights and pretrained strategy are in Appendix \ref{sec:source_pretraining}.

For test-time adaptation via fine-tuning, the student model on CIFAR-10 is trained for $30$ epochs using SGD with a decaying learning-rate schedule. The learning rate is initialized at $10^{-3}$ and decayed at epochs $10$, $25$, and $30$. During adversarial fine-tuning, adversarial examples are generated according to each specific optimization strategy, using perturbation budget $\varepsilon={8}/{255}$, step size $\alpha={2}/{255}$, and $5$ attack steps. Unless otherwise stated, the same adversarial hyperparameters are used across all compared adversarial training strategies, with $\beta=6$.

For ImageNet with ResNet-50, we follow the training setup of~\cite{wong2020fast,shafahi2019adversarial}, performing adversarial fine-tuning for $30$ epochs with SGD and a decaying learning-rate schedule. We generate adversarial examples using $5$ attack steps and step size $\alpha={1}/{255}$, with learning rate $10^{-3}$. We consider two perturbation budgets, $\varepsilon \in \{{2}/{255}, {4}/{255}\}$.
For ImageNet with ViT-B/16, we follow the setup of~\cite{mo2022adversarial}. In this case, we use $5$ attack steps, perturbation budget $\varepsilon={2}/{255}$, step size $\alpha={1}/{255}$, and learning rate $10^{-4}$.

For the teacher, we adopt a simple batch-normalization-based adaptation scheme, where batch normalization layers are kept active during test-time adaptation~\cite{li2016revisiting}. This allows the teacher to better align with the statistics of the corrupted target batches, improving the quality of its clean predictions, which is particularly beneficial in our setting, as the teacher predictions are used to guide clean and robust objectives.

\medskip
\noindent \textbf{Metrics and attacks.}
We report both clean and robust accuracy. Clean accuracy is measured on clean target samples, whereas robust accuracy is defined as the fraction of adversarially perturbed inputs that remain correctly classified. Adversarial robustness is evaluated under an $L_\infty$ threat model using multiple attacks, including $20$-step PGD \cite{madry2017towards}, PGN~\cite{ge2023boosting}, and Square Attack~\cite{andriushchenko2020square}. For CIFAR-10, we use $\varepsilon={8}/{255}$ and step size $\alpha={2}/{255}$ during evaluation. For ImageNet, evaluation is performed using the same perturbation budget adopted during adaptation.


\subsection{Evaluation of Test-Time Adaptation}
This section evaluates the performance of \methodname{} compared with \othermethod{} for unsupervised test-time robust adaptation, across different $\beta$ configurations.

\paragraph{Evaluation on CIFAR-10.}
\begin{table}[t]
\centering
\caption{\small{Clean and robust accuracy on the {CIFAR-10} test set for unsupervised robust test-time adaptation on CIFAR-10 training set, evaluated across different values of $\beta$ for \methodname and \othermethod under varying target-domain severity levels (for both training and testing).}}
\label{tab:full_cifar10}
\resizebox{0.48\textwidth}{!}{
\begin{tabular}{c cccc cccc}
\toprule
\multirow{2}{*}{$\beta$} 
& \multicolumn{4}{c }{\methodname (Ours)} 
& \multicolumn{4}{c}{\othermethod} \\
\cline{2-9}
& \cellcolor{gray!20}Clean & PGD & PGN & Square
& \cellcolor{gray!20}Clean & PGD & PGN & Square \\
\midrule

\multicolumn{9}{c}{\textbf{Target Domain Severity = 0} (No shift, i.e., robust fine-tuning)} \\
\midrule
6  & \cellcolor{gray!20}88.22 & 46.39 & 54.77 & 53.18 & \cellcolor{gray!20}79.06 & 45.50 & 49.69 & 46.94 \\
8  & \cellcolor{gray!20}87.95 & 46.87 & 55.40 & 53.29 & \cellcolor{gray!20}76.12 & 44.93 & 48.31 & 45.48 \\
10 & \cellcolor{gray!20}88.11 & 47.30 & 55.42 & 53.63 & \cellcolor{gray!20}73.49 & 43.67 & 46.72 & 43.67 \\
12 & \cellcolor{gray!20}87.89 & 47.47 & 55.63 & 53.55 & \cellcolor{gray!20}66.65 & 39.32 & 41.13 & 38.52 \\
\midrule

\multicolumn{9}{c}{\textbf{Target Domain Severity = 1}} \\
\midrule
6  & \cellcolor{gray!20}78.19 & 40.23 & 46.35 & 45.55 & \cellcolor{gray!20}71.66 & 36.02 & 39.96 & 38.62 \\
8  & \cellcolor{gray!20}77.86 & 40.76 & 47.06 & 46.31 & \cellcolor{gray!20}67.27 & 35.69 & 38.09 & 37.35 \\
10 & \cellcolor{gray!20}77.73 & 40.81 & 47.27 & 46.61 & \cellcolor{gray!20}64.29 & 34.55 & 36.73 & 35.97 \\
12 & \cellcolor{gray!20}77.50 & 41.13 & 47.31 & 46.20 & \cellcolor{gray!20}61.92 & 34.94 & 36.47 & 34.92 \\
\midrule

\multicolumn{9}{c}{\textbf{Target Domain Severity = 2}} \\
\midrule
6  & \cellcolor{gray!20}64.19 & 33.18 & 37.75 & 35.45 & \cellcolor{gray!20}56.64 & 26.74 & 28.59 & 27.85 \\
8  & \cellcolor{gray!20}64.05 & 33.40 & 38.32 & 36.39 & \cellcolor{gray!20}53.18 & 25.77 & 27.95 & 26.42 \\
10 & \cellcolor{gray!20}63.34 & 34.06 & 38.59 & 36.51 & \cellcolor{gray!20}49.62 & 25.23 & 26.02 & 24.89 \\
12 & \cellcolor{gray!20}63.16 & 34.18 & 39.19 & 36.17 & \cellcolor{gray!20}46.70 & 23.28 & 23.95 & 23.70 \\
\bottomrule
\end{tabular}
}
\end{table}

We first focus on CIFAR-10, considering the setting in which the target domain dataset $D_T$ is derived from the CIFAR-10 training set.
Table~\ref{tab:full_cifar10} shows a consistent advantage of \methodname{} over \othermethod{} across all corruption severities and all values of $\beta$, with a significantly better robustness-accuracy trade-off as the corruption severity increases from $0$ to $2$.
Importantly, the performance gap tends to widen as the shift becomes stronger, indicating that the teacher-anchored objective is particularly beneficial under larger domain shifts, consistently with the analysis discussed in Section \ref{sec:theory_stability}. For instance, at severity $2$ and $\beta=12$, \methodname{} improves over \othermethod{} by more than $16$ points in clean accuracy and by about $11$ points under PGD, with similarly large gains under PGN and Square Attack. This trend confirms that anchoring both the clean and adversarial objectives to the teacher provides a more stable adaptation signal than relying on the student predictions alone.

Another important result is that \methodname{} is considerably less sensitive to $\beta$ than \othermethod{}. As $\beta$ increases, \methodname{} exhibits only minor variations in clean accuracy and often a small but consistent gain in robust accuracy, whereas \othermethod{} suffers a pronounced degradation in clean performance with only limited robustness improvement. This effect is even more evident under a stronger distribution shift, where increasing $\beta$ in \othermethod{} leads to a clear collapse in clean accuracy, while \methodname{} remains stable. 

\paragraph{Evaluation on ImageNet.}
\begin{table*}[t]
\centering
\caption{\small{Clean and robust accuracy on the ResNet-50 ImageNet validation set across different values of $\beta$ for \methodname{} and \othermethod{} under varying target-domain severity levels.}}
\label{tab:resnet_combined}
\begin{subtable}[t]{0.49\textwidth}
\centering
\caption{$\varepsilon=2$.}
\label{tab:resnet_eps_2}
\resizebox{\textwidth}{!}{
\begin{tabular}{c cc cc cc cc}
\toprule
\multirow{3}{*}{$\beta$} 
& \multicolumn{4}{c }{\methodname{} (Ours)} 
& \multicolumn{4}{c}{\othermethod{}} \\
\cline{2-9}
& \multicolumn{2}{c}{Clean} & \multicolumn{2}{c }{Robust}
& \multicolumn{2}{c}{Clean} & \multicolumn{2}{c}{Robust} \\
\cline{2-9}
& \cellcolor{gray!20}T1 & \cellcolor{gray!20}T5 & T1 & T5
& \cellcolor{gray!20}T1 & \cellcolor{gray!20}T5 & T1 & T5 \\
\midrule
\multicolumn{9}{c}{\textbf{Target Domain Severity = 0}} \\
\midrule
2  & \cellcolor{gray!20}67.94 & \cellcolor{gray!20}88.10 & 30.86 & 63.26 & \cellcolor{gray!20}68.64 & \cellcolor{gray!20}88.36 & 25.38 & 56.74 \\
4  & \cellcolor{gray!20}65.61 & \cellcolor{gray!20}86.51 & 32.34 & 63.57 & \cellcolor{gray!20}66.23 & \cellcolor{gray!20}86.95 & 29.07 & 60.22 \\
6  & \cellcolor{gray!20}64.13 & \cellcolor{gray!20}85.62 & 33.88 & 64.04 & \cellcolor{gray!20}64.03 & \cellcolor{gray!20}85.48 & 29.60 & 60.14 \\
8  & \cellcolor{gray!20}63.37 & \cellcolor{gray!20}85.07 & 34.28 & 64.00 & \cellcolor{gray!20}55.52 & \cellcolor{gray!20}78.30 & 31.49 & 58.06 \\
10 & \cellcolor{gray!20}63.11 & \cellcolor{gray!20}84.77 & 34.48 & 64.01 & \cellcolor{gray!20}54.48 & \cellcolor{gray!20}77.54 & 31.07 & 57.09 \\
12 & \cellcolor{gray!20}62.50 & \cellcolor{gray!20}84.42 & 34.47 & 63.88 & \cellcolor{gray!20}53.12 & \cellcolor{gray!20}76.34 & 30.89 & 56.47 \\
\midrule
\multicolumn{9}{c}{\textbf{Target Domain Severity = 1}} \\
\midrule
2  & \cellcolor{gray!20}62.83 & \cellcolor{gray!20}84.07 & 25.64 & 54.04 & \cellcolor{gray!20}63.95 & \cellcolor{gray!20}84.98 & 20.48 & 46.40 \\
4  & \cellcolor{gray!20}59.74 & \cellcolor{gray!20}81.98 & 27.40 & 55.26 & \cellcolor{gray!20}55.54 & \cellcolor{gray!20}78.66 & 26.34 & 51.92 \\
6  & \cellcolor{gray!20}58.94 & \cellcolor{gray!20}81.42 & 27.66 & 55.51 & \cellcolor{gray!20}58.22 & \cellcolor{gray!20}81.08 & 24.51 & 51.44 \\
8  & \cellcolor{gray!20}57.38 & \cellcolor{gray!20}80.02 & 28.33 & 55.32 & \cellcolor{gray!20}55.83 & \cellcolor{gray!20}79.26 & 24.74 & 50.71 \\
10 & \cellcolor{gray!20}57.06 & \cellcolor{gray!20}79.91 & 28.60 & 55.64 & \cellcolor{gray!20}52.84 & \cellcolor{gray!20}76.83 & 24.54 & 50.00 \\
12 & \cellcolor{gray!20}55.73 & \cellcolor{gray!20}78.92 & 28.96 & 55.46 & \cellcolor{gray!20}47.01 & \cellcolor{gray!20}70.93 & 24.03 & 47.73 \\
\midrule
\multicolumn{9}{c}{\textbf{Target Domain Severity = 2}} \\
\midrule
2  & \cellcolor{gray!20}59.19 & \cellcolor{gray!20}81.71 & 21.22 & 46.14 & \cellcolor{gray!20}59.99 & \cellcolor{gray!20}82.37 & 16.01 & 35.65 \\
4  & \cellcolor{gray!20}56.14 & \cellcolor{gray!20}78.85 & 23.52 & 48.42 & \cellcolor{gray!20}57.03 & \cellcolor{gray!20}79.73 & 19.62 & 43.13 \\
6  & \cellcolor{gray!20}53.76 & \cellcolor{gray!20}77.37 & 24.42 & 49.47 & \cellcolor{gray!20}47.69 & \cellcolor{gray!20}71.74 & 22.50 & 44.22 \\
8  & \cellcolor{gray!20}52.69 & \cellcolor{gray!20}76.34 & 24.94 & 49.70 & \cellcolor{gray!20}45.77 & \cellcolor{gray!20}69.81 & 22.27 & 43.80 \\
10 & \cellcolor{gray!20}51.86 & \cellcolor{gray!20}75.62 & 25.15 & 49.94 & \cellcolor{gray!20}43.79 & \cellcolor{gray!20}68.02 & 20.46 & 42.15 \\
12 & \cellcolor{gray!20}51.34 & \cellcolor{gray!20}75.31 & 25.34 & 49.74 & \cellcolor{gray!20}46.14 & \cellcolor{gray!20}70.40 & 20.46 & 42.62 \\
\bottomrule
\end{tabular}
}
\end{subtable}
\hfill
\begin{subtable}[t]{0.49\textwidth}
\centering
\caption{$\varepsilon=4$.}
\label{tab:resnet_eps_4}
\resizebox{\textwidth}{!}{
\begin{tabular}{c cc cc cc cc}
\toprule
\multirow{3}{*}{$\beta$} 
& \multicolumn{4}{c }{\methodname{} (Ours)} 
& \multicolumn{4}{c}{\othermethod{}} \\
\cline{2-9}
& \multicolumn{2}{c }{Clean} & \multicolumn{2}{c }{Robust}
& \multicolumn{2}{c }{Clean} & \multicolumn{2}{c}{Robust} \\
\cline{2-9}
& \cellcolor{gray!20}T1 & \cellcolor{gray!20}T5 & T1 & T5
& \cellcolor{gray!20}T1 & \cellcolor{gray!20}T5 & T1 & T5 \\
\midrule
\multicolumn{9}{c}{\textbf{Target Domain Severity = 0}} \\
\midrule
2  & \cellcolor{gray!20}60.94 & \cellcolor{gray!20}82.78 & 12.97 & 32.84 & \cellcolor{gray!20}63.94 & \cellcolor{gray!20}85.20 & 13.06 & 32.35 \\
4  & \cellcolor{gray!20}56.89 & \cellcolor{gray!20}79.96 & 16.06 & 40.60 & \cellcolor{gray!20}59.72 & \cellcolor{gray!20}82.28 & 13.14 & 34.56 \\
6  & \cellcolor{gray!20}56.13 & \cellcolor{gray!20}79.14 & 16.30 & 40.40 & \cellcolor{gray!20}57.71 & \cellcolor{gray!20}80.74 & 13.03 & 35.80 \\
8  & \cellcolor{gray!20}52.75 & \cellcolor{gray!20}76.72 & 16.71 & 40.00 & \cellcolor{gray!20}56.17 & \cellcolor{gray!20}79.35 & 13.42 & 36.13 \\
10 & \cellcolor{gray!20}52.08 & \cellcolor{gray!20}76.08 & 16.96 & 40.03 & \cellcolor{gray!20}55.10 & \cellcolor{gray!20}78.77 & 13.77 & 36.79 \\
12 & \cellcolor{gray!20}51.27 & \cellcolor{gray!20}75.32 & 17.24 & 39.94 & \cellcolor{gray!20}42.80 & \cellcolor{gray!20}66.24 & 16.52 & 35.19 \\
\midrule
\multicolumn{9}{c}{\textbf{Target Domain Severity = 1}} \\
\midrule
2  & \cellcolor{gray!20}55.22 & \cellcolor{gray!20}78.11 & 11.26 & 29.47 & \cellcolor{gray!20}58.22 & \cellcolor{gray!20}80.75 & 9.65 & 23.30 \\
4  & \cellcolor{gray!20}53.25 & \cellcolor{gray!20}76.34 & 12.77 & 32.36 & \cellcolor{gray!20}55.38 & \cellcolor{gray!20}78.64 & 10.78 & 26.32 \\
6  & \cellcolor{gray!20}48.08 & \cellcolor{gray!20}71.82 & 13.22 & 32.73 & \cellcolor{gray!20}44.92 & \cellcolor{gray!20}68.54 & 12.98 & 29.56 \\
8  & \cellcolor{gray!20}46.87 & \cellcolor{gray!20}71.11 & 13.48 & 32.62 & \cellcolor{gray!20}52.02 & \cellcolor{gray!20}75.89 & 11.11 & 28.72 \\
10 & \cellcolor{gray!20}45.64 & \cellcolor{gray!20}69.98 & 13.70 & 32.28 & \cellcolor{gray!20}50.94 & \cellcolor{gray!20}75.02 & 10.98 & 28.60 \\
12 & \cellcolor{gray!20}45.44 & \cellcolor{gray!20}69.82 & 14.02 & 32.68 & \cellcolor{gray!20}39.90 & \cellcolor{gray!20}63.67 & 14.10 & 30.78 \\
\midrule
\multicolumn{9}{c}{\textbf{Target Domain Severity = 2}} \\
\midrule
2  & \cellcolor{gray!20}56.33 & \cellcolor{gray!20}78.66 & 3.96 & 9.08 & \cellcolor{gray!20}55.35 & \cellcolor{gray!20}78.16 & 7.17 & 15.56 \\
4  & \cellcolor{gray!20}50.50 & \cellcolor{gray!20}74.42 & 10.48 & 26.28 & \cellcolor{gray!20}51.85 & \cellcolor{gray!20}75.45 & 8.78 & 20.63 \\
6  & \cellcolor{gray!20}45.79 & \cellcolor{gray!20}69.49 & 11.54 & 28.44 & \cellcolor{gray!20}44.77 & \cellcolor{gray!20}68.74 & 8.78 & 22.28 \\
8  & \cellcolor{gray!20}44.12 & \cellcolor{gray!20}68.30 & 12.10 & 29.22 & \cellcolor{gray!20}49.14 & \cellcolor{gray!20}73.30 & 8.96 & 22.88 \\
10 & \cellcolor{gray!20}43.39 & \cellcolor{gray!20}67.76 & 12.14 & 29.22 & \cellcolor{gray!20}47.25 & \cellcolor{gray!20}71.75 & 9.49 & 23.89 \\
12 & \cellcolor{gray!20}42.20 & \cellcolor{gray!20}66.46 & 12.33 & 28.93 & \cellcolor{gray!20}36.28 & \cellcolor{gray!20}59.49 & 10.40 & 23.85 \\
\bottomrule
\end{tabular}
}
\end{subtable}
\end{table*}

\begin{table}[t]
\centering
\caption{\small{Clean and robust accuracy on the ViT-B ImageNet validation set across different values of $\beta$ for \methodname{} and \othermethod{} under varying target-domain severity levels. For attacks $\varepsilon=2$}.}
\label{tab:vit}
\resizebox{0.48\textwidth}{!}{
\begin{tabular}{c cc cc cc cc}
\toprule
\multirow{3}{*}{$\beta$} 
& \multicolumn{4}{c}{\methodname{} (Ours)} 
& \multicolumn{4}{c}{\othermethod{}} \\
\cline{2-9}
& \multicolumn{2}{c}{Clean} & \multicolumn{2}{c}{Robust}
& \multicolumn{2}{c}{Clean} & \multicolumn{2}{c}{Robust} \\
\cline{2-9}
& \cellcolor{gray!20}T1 & \cellcolor{gray!20}T5 & T1 & T5
& \cellcolor{gray!20}T1 & \cellcolor{gray!20}T5 & T1 & T5 \\
\midrule

\multicolumn{9}{c}{\textbf{Target Domain Severity = 0}} \\
\midrule
6  & \cellcolor{gray!20}71.92 & \cellcolor{gray!20}90.29 & 39.34 & 73.99 & \cellcolor{gray!20}67.03 & \cellcolor{gray!20}87.10 & 40.08 & 69.70 \\
8  & \cellcolor{gray!20}71.70 & \cellcolor{gray!20}90.06 & 39.62 & 74.11 & \cellcolor{gray!20}66.01 & \cellcolor{gray!20}86.30 & 40.37 & 69.33 \\
10 & \cellcolor{gray!20}71.42 & \cellcolor{gray!20}89.85 & 39.80 & 74.14 & \cellcolor{gray!20}63.76 & \cellcolor{gray!20}84.81 & 38.80 & 67.36 \\
12 & \cellcolor{gray!20}71.04 & \cellcolor{gray!20}88.12 & 39.50 & 74.10 & \cellcolor{gray!20}61.45 & \cellcolor{gray!20}82.32 & 37.10 & 66.56 \\
\midrule

\multicolumn{9}{c}{\textbf{Target Domain Severity = 1}} \\
\midrule
6  & \cellcolor{gray!20}68.34 & \cellcolor{gray!20}87.96 & 35.46 & 68.97 & \cellcolor{gray!20}62.67 & \cellcolor{gray!20}83.86 & 34.78 & 63.43 \\
8  & \cellcolor{gray!20}68.07 & \cellcolor{gray!20}87.72 & 35.50 & 68.96 & \cellcolor{gray!20}61.18 & \cellcolor{gray!20}82.62 & 34.97 & 62.93 \\
10 & \cellcolor{gray!20}67.72 & \cellcolor{gray!20}87.42 & 35.70 & 69.02 & \cellcolor{gray!20}59.86 & \cellcolor{gray!20}81.62 & 34.87 & 62.31 \\
12 & \cellcolor{gray!20}68.01 & \cellcolor{gray!20}87.32 & 35.60 & 68.57 & \cellcolor{gray!20}58.46 & \cellcolor{gray!20}80.05 & 34.56 & 62.67 \\
\midrule

\multicolumn{9}{c}{\textbf{Target Domain Severity = 2}} \\
\midrule
6  & \cellcolor{gray!20}64.90 & \cellcolor{gray!20}85.42 & 32.00 & 63.98 & \cellcolor{gray!20}58.03 & \cellcolor{gray!20}80.14 & 30.26 & 57.15 \\
8  & \cellcolor{gray!20}64.43 & \cellcolor{gray!20}85.22 & 32.38 & 64.06 & \cellcolor{gray!20}56.26 & \cellcolor{gray!20}78.92 & 30.49 & 56.87 \\
10 & \cellcolor{gray!20}64.24 & \cellcolor{gray!20}85.06 & 32.71 & 64.16 & \cellcolor{gray!20}54.74 & \cellcolor{gray!20}77.55 & 30.24 & 55.88 \\
12 & \cellcolor{gray!20}64.13 & \cellcolor{gray!20}85.00 & 32.65 & 64.23 & \cellcolor{gray!20}53.25 & \cellcolor{gray!20}76.68 & 30.14 & 55.32 \\
\bottomrule
\end{tabular}
}
\end{table}
For ImageNet, we evaluate robustness using PGD, as in prior work \cite{croce2020robustbench,croce2020reliable}, and report both Top-1 and Top-5 accuracy. In particular, Table~\ref{tab:resnet_eps_2} and Table~\ref{tab:resnet_eps_4} report the results for ResNet-50 under adversarial fine-tuning budgets of $\varepsilon=2/255$ and $\varepsilon=4/255$, respectively, while Table~\ref{tab:vit} reports the results for ViT-B/16.

Overall, \methodname{} consistently provides a better robustness-accuracy trade-off than \othermethod{}. For ResNet-50 with $\varepsilon=2/255$ (Table~\ref{tab:resnet_eps_2}), the improvement is systematic across all corruption severities and becomes more evident as $\beta$ increases: \methodname{} preserves comparable or higher clean accuracy while achieving stronger PGD robustness, and is noticeably less sensitive to the choice of $\beta$. When the adversarial budget is increased to $\varepsilon=4/255$ (Table~\ref{tab:resnet_eps_4}), the setting becomes more challenging and the clean-accuracy comparison is less uniform; nevertheless, \methodname{} still provides better robust performance in almost all cases, especially at moderate $\beta$ and under stronger shift. The results on ViT-B/16 (Table~\ref{tab:vit}) follow a similar trend: \methodname{} consistently outperforms \othermethod{} in clean accuracy and in robust Top-5 accuracy across all severities, while also improving robust Top-1 accuracy in nearly all settings.

Overall, both results on CIFAR10 and ImageNet Val indicate that anchoring both the clean and adversarial objectives to a fixed teacher yields a more stable adaptation mechanism across architectures, corruption levels, and target-domain settings, and further support the theoretical analysis discussed in Section \ref{sec:theory_stability}.

\begin{figure*}[h]
    \centering
    \begin{subfigure}{0.8\textwidth}
        \centering
        \includegraphics[width=\textwidth]{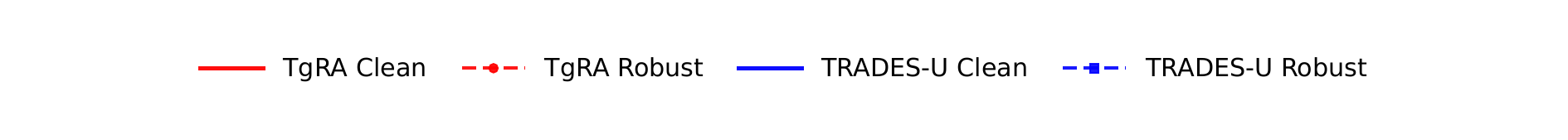}
    \end{subfigure}
    \begin{subfigure}{0.95\textwidth}
        \centering
        \includegraphics[width=\textwidth]{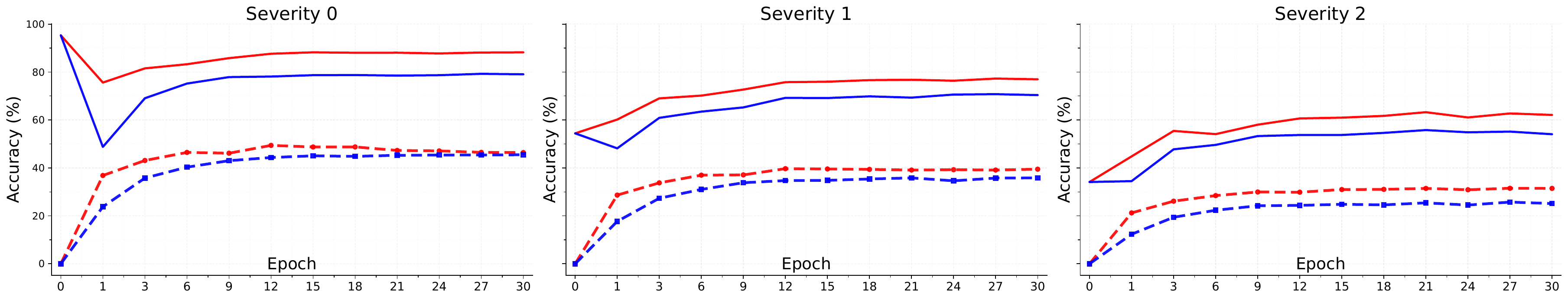}
        \caption{CIFAR10 - WideResNet-34}
        \label{f:dynamics_analysis_cifar10}
    \end{subfigure}
    \begin{subfigure}{0.95\textwidth}
        \centering
        \includegraphics[width=\textwidth]{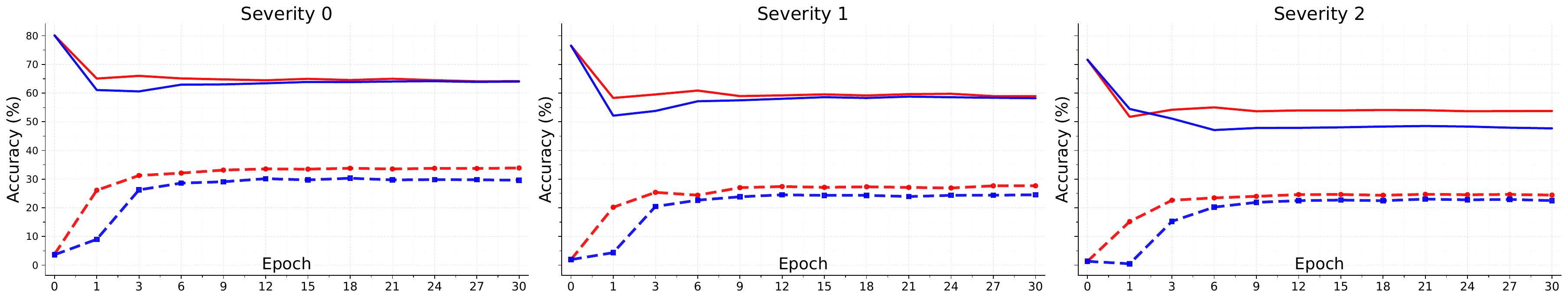}
        \caption{ImageNetVal - ResNet50}
        \label{f:dynamics_analysis_imagenet}
    \end{subfigure}
    \caption{\small{Clean and robust test accuracy under PGD-20 attacks during test-time finetuning on CIFAR-10 ($\varepsilon = 8/255$) and ImageNet-Val ($\varepsilon = 2/255$), evaluated across different levels of distribution-shift severity. }}
    \label{f:dynamics_analysis}
\end{figure*}



\subsection{Analysis of the training dynamics}
We analyze in Figure~\ref{f:dynamics_analysis} the evolution of clean and robust accuracy during unsupervised adversarial test-time finetuning, where robustness is evaluated under a PGD-20 attack across increasing levels of distribution shift.

For CIFAR-10 (Figure~\ref{f:dynamics_analysis_cifar10}), \methodname{} exhibits a more stable and consistent improvement in both clean and robust accuracy throughout training compared with \othermethod{}. This behavior becomes particularly evident when the shift is non-negligible ($\text{severity} > 0$), where \methodname{} consistently achieves higher robust accuracy and avoids the pronounced degradation in clean performance observed in competing methods.
Note also that the improvements in clean accuracy at severity levels greater than 0 can be explained by the parallel adaptation of the teacher model. Specifically, while the student is finetuned, the teacher also undergoes test-time adaptation through batch normalization updates, allowing it to better match the shifted distribution. This adapted teacher can then distill more reliable knowledge to the student, ultimately improving the student’s clean accuracy.

For ImageNet (Figure~\ref{f:dynamics_analysis_imagenet}), the plots show a similar trend across all severity levels, with \methodname{} exhibiting the most favorable convergence behavior. After the initial drop in clean accuracy during the first epochs, it recovers quickly, stabilizes after a few iterations, and consistently achieves the best robustness--accuracy trade-off. Also in this case, the advantage becomes more evident as the shift becomes more pronounced, namely for severity levels greater than 0, confirming that the teacher-guided objective provides a more stable adaptation signal under limited unlabeled target data, especially when the distribution shift is stronger.

A key observation across all analyses is the large accuracy gap that emerges in the early epochs for \othermethod{}. In \othermethod{}, the robustness term is mainly enforced through self-consistency constraints based on the student’s own predictions. However, since the student is not robust a priori and, at the beginning, has a confidence similar to that of the teacher on the shifted target domain, this self-referential signal may generate unstable or misleading gradients, causing an early drop in clean accuracy and a slower recovery during training (see the theoretical analysis in Section~\ref{sec:theory_stability}). 

\subsection{Importance of target samples}
\label{sec:target_samples}
As is well known in the literature, data availability plays a crucial role in classical adversarial training \cite{schmidt2018adversarially,wang2020tent,zhao2023pitfalls}, and consequently also in finetuning procedures. This aspect becomes even more critical in the test-time regime, where data are typically not only unlabeled but also scarce. For this reason, we study the impact of data availability in this setting and analyze how the considered techniques behave under such constraints. 

\begin{table}[t]
\centering
\caption{\small{Clean and robust Accuracy on the {split 50/50 CIFAR-10 test set} for test-time unsupervised robust adaptation, evaluated across different values of $\beta$ for \methodname{} and \othermethod{} under varying target-domain severity levels. For clean and PGD robustness, the value in parentheses denotes the gap with respect to the corresponding result obtained in Table\ref{tab:full_cifar10}.}}
\label{tab:split_cifar10}
\resizebox{0.5\textwidth}{!}{
\begin{tabular}{c cc cc}
\toprule
\multirow{2}{*}{$\beta$} 
& \multicolumn{2}{c}{\methodname{} (Ours)} 
& \multicolumn{2}{c}{\othermethod{}} \\
\cmidrule(lr){2-3} \cmidrule(lr){4-5}
& \cellcolor{gray!20}Clean & Robust 
& \cellcolor{gray!20}Clean & Robust \\
\midrule

\multicolumn{5}{c}{\textbf{Target Domain Severity = 0}} \\
\midrule
6  & \cellcolor{gray!20}73.46 {\scriptsize($-14.76$)} & 28.54 {\scriptsize($-17.85$)} & \cellcolor{gray!20}66.42 {\scriptsize($-12.64$)} & 13.70 {\scriptsize($-31.80$)} \\
8  & \cellcolor{gray!20}73.38 {\scriptsize($-14.57$)} & 30.28 {\scriptsize($-16.59$)} & \cellcolor{gray!20}58.92 {\scriptsize($-17.20$)} & 14.46 {\scriptsize($-30.47$)} \\
10 & \cellcolor{gray!20}73.78 {\scriptsize($-14.33$)} & 31.40 {\scriptsize($-15.90$)} & \cellcolor{gray!20}51.30 {\scriptsize($-22.19$)} & 15.02 {\scriptsize($-28.65$)} \\
12 & \cellcolor{gray!20}74.20 {\scriptsize($-13.69$)} & 32.24 {\scriptsize($-15.23$)} & \cellcolor{gray!20}47.72 {\scriptsize($-18.93$)} & 16.10 {\scriptsize($-23.22$)} \\
\midrule

\multicolumn{5}{c}{\textbf{Target Domain Severity = 1}} \\
\midrule
6  & \cellcolor{gray!20}57.14 {\scriptsize($-21.05$)} & 21.76 {\scriptsize($-18.47$)} & \cellcolor{gray!20}54.36 {\scriptsize($-17.30$)} & 4.68 {\scriptsize($-31.34$)} \\
8  & \cellcolor{gray!20}56.06 {\scriptsize($-21.80$)} & 23.22 {\scriptsize($-17.54$)} & \cellcolor{gray!20}44.44 {\scriptsize($-22.83$)} & 5.58 {\scriptsize($-30.11$)} \\
10 & \cellcolor{gray!20}55.60 {\scriptsize($-22.13$)} & 25.14 {\scriptsize($-15.67$)} & \cellcolor{gray!20}37.54 {\scriptsize($-26.75$)} & 6.84 {\scriptsize($-27.71$)} \\
12 & \cellcolor{gray!20}56.36 {\scriptsize($-21.14$)} & 25.64 {\scriptsize($-15.49$)} & \cellcolor{gray!20}33.96 {\scriptsize($-27.96$)} & 7.70 {\scriptsize($-27.24$)} \\
\midrule

\multicolumn{5}{c}{\textbf{Target Domain Severity = 2}} \\
\midrule
6  & \cellcolor{gray!20}42.70 {\scriptsize($-21.49$)} & 16.30 {\scriptsize($-16.88$)} & \cellcolor{gray!20}31.96 {\scriptsize($-24.68$)} & 2.70 {\scriptsize($-24.04$)} \\
8  & \cellcolor{gray!20}42.66 {\scriptsize($-21.39$)} & 18.48 {\scriptsize($-14.92$)} & \cellcolor{gray!20}26.64 {\scriptsize($-26.54$)} & 3.50 {\scriptsize($-22.27$)} \\
10 & \cellcolor{gray!20}42.94 {\scriptsize($-20.40$)} & 19.02 {\scriptsize($-15.04$)} & \cellcolor{gray!20}25.14 {\scriptsize($-24.48$)} & 2.98 {\scriptsize($-22.25$)} \\
12 & \cellcolor{gray!20}43.58 {\scriptsize($-19.58$)} & 20.30 {\scriptsize($-13.88$)} & \cellcolor{gray!20}26.30 {\scriptsize($-20.40$)} & 4.50 {\scriptsize($-18.78$)} \\
\bottomrule
\end{tabular}
}
\end{table}

We consider a challenging setting for CIFAR-10 in which the target domain is built from a corrupted 50/50 split of the CIFAR-10 test set, while the source model is still pretrained on the CIFAR-10 training set, as in the previous analysis.
Table~\ref{tab:split_cifar10} reports the results of both \methodname{} and \othermethod{} across all values of $\beta$ and all corruption severities. A clear performance gap emerges with respect to the use of larger datasets, as in Table~\ref{tab:full_cifar10}, which is better highlighted by the values reported in parentheses. This trend is consistent with the well-known importance of additional data for adversarial training. Despite this, we observe that such a degradation is significantly less pronounced for \methodname{} than for \othermethod{}, with the latter becoming increasingly unstable as $\beta$ grows and suffering a marked drop in clean accuracy without comparable gains in robustness. This advantage is already visible in the no-shift case and becomes even more pronounced under distribution shift.

\subsection{Comparisons with supervised approaches}
\label{sec:supervised}
To conclude, we also analyze in Table~\ref{tab:supervised_comparison_all} the behavior of unsupervised robust test-time adaptation in comparison with a supervised scenario in which target labels are available during adaptation. This setting is mainly addressed by well-known adversarial training techniques, which are described and compared below.

For the supervised finetuning baselines, we consider PGD~\cite{madry2017towards}, TRADES~\cite{zhang2019theoretically}, MART~\cite{wang2019improving}, and DKL~\cite{cui2024decoupled}. In particular, for TRADES and MART we use $\beta = 6$, while for DKL, following the original paper, we use $\beta = 20$ with an initial learning rate of $1.5 \cdot 10^{-3}$ and a cosine-annealing learning-rate scheduler. These methods use target labels during adaptation and therefore do not belong to the same unsupervised setting as \methodname{} and \othermethod{}, which are separated in the tables for clarity.

\begin{table*}[t]
\centering
\caption{\small{Comparison between \methodname{} and supervised adversarial finetuning approaches, evaluated in terms of clean and PGD-20 robust accuracy under varying target-domain severity levels (for both training and testing).}}
\label{tab:supervised_comparison_all}
\vspace{2mm}
\resizebox{0.75\textwidth}{!}{
\begin{subtable}[t]{0.26\textwidth}
\centering
\caption{\small{WRNet-34 on CIFAR-10}}
\label{tab:full_cifar10_supervised}
\resizebox{\textwidth}{!}{%
\begin{tabular}{lcc}
\toprule
Method & \cellcolor{gray!20} Clean & Robust \\
\midrule
\multicolumn{3}{c}{\textbf{Severity = 0}} \\
\midrule
PGD           & \cellcolor{gray!20}86.83 & 43.76 \\
TRADES        & \cellcolor{gray!20}79.22 & 45.68 \\
MART          & \cellcolor{gray!20}93.91 & 25.08 \\
DKL           & \cellcolor{gray!20}76.73 & 40.69 \\
\hdashline
\othermethod  & \cellcolor{gray!20}79.06 & 45.48 \\
\methodname   & \cellcolor{gray!20}88.22 & 46.39 \\
\midrule
\multicolumn{3}{c}{\textbf{Severity = 1}} \\
\midrule
PGD           & \cellcolor{gray!20}73.70 & 35.92 \\
TRADES        & \cellcolor{gray!20}58.93 & 31.30 \\
MART          & \cellcolor{gray!20}82.27 & 0.00 \\
DKL           & \cellcolor{gray!20}65.32 & 32.65 \\
\hdashline
\othermethod  & \cellcolor{gray!20}70.38 & 35.82 \\
\methodname   & \cellcolor{gray!20}76.97 & 39.46 \\
\midrule
\multicolumn{3}{c}{\textbf{Severity = 2}} \\
\midrule
PGD           & \cellcolor{gray!20}66.51 & 32.23 \\
TRADES        & \cellcolor{gray!20}48.43 & 25.92 \\
MART          & \cellcolor{gray!20}30.69 & 16.51 \\
DKL           & \cellcolor{gray!20}60.04 & 29.19 \\
\hdashline
\othermethod  & \cellcolor{gray!20}54.05 & 25.13 \\
\methodname   & \cellcolor{gray!20}62.10 & 31.45 \\
\bottomrule
\end{tabular}}
\end{subtable}
\hfill
\begin{subtable}[t]{0.26\textwidth}
\centering
\caption{\small{WRNet-34, CIFAR-10-Test (50)}}
\label{tab:split_cifar10_supervised}
\resizebox{\textwidth}{!}{%
\begin{tabular}{lcc}
\toprule
Method & \cellcolor{gray!20} Clean & Robust \\
\midrule
\multicolumn{3}{c}{\textbf{Severity = 0}} \\
\midrule
PGD           & \cellcolor{gray!20}70.52 & 16.84 \\
TRADES        & \cellcolor{gray!20}60.28 & 11.58 \\
MART          & \cellcolor{gray!20}89.00 & 9.48 \\
DKL           & \cellcolor{gray!20}66.14 & 25.54 \\
\hdashline
\othermethod  & \cellcolor{gray!20}67.82 & 15.68 \\
\methodname   & \cellcolor{gray!20}74.38 & 32.28 \\
\midrule
\multicolumn{3}{c}{\textbf{Severity = 1}} \\
\midrule
PGD           & \cellcolor{gray!20}54.94 & 3.88 \\
TRADES        & \cellcolor{gray!20}42.22 & 6.80 \\
MART          & \cellcolor{gray!20}53.54 & 4.50 \\
DKL           & \cellcolor{gray!20}56.68 & 19.18 \\
\hdashline
\othermethod  & \cellcolor{gray!20}54.14 & 4.52 \\
\methodname   & \cellcolor{gray!20}56.36 & 25.64 \\
\midrule
\multicolumn{3}{c}{\textbf{Severity = 2}} \\
\midrule
PGD           & \cellcolor{gray!20}37.10 & 1.84 \\
TRADES        & \cellcolor{gray!20}33.62 & 9.18 \\
MART          & \cellcolor{gray!20}36.34 & 7.74 \\
DKL           & \cellcolor{gray!20}50.76 & 16.20 \\
\hdashline
\othermethod  & \cellcolor{gray!20}26.30 & 4.50 \\
\methodname   & \cellcolor{gray!20}43.58 & 20.30 \\
\bottomrule
\end{tabular}}
\end{subtable}
\hfill
\begin{subtable}[t]{0.26\textwidth}
\centering
\caption{\small{ResNet-50 on ImageNet-Val}}
\label{tab:imagenet_supervised}
\resizebox{\textwidth}{!}{%
\begin{tabular}{lcc}
\toprule
Method & \cellcolor{gray!20} Clean & Robust \\
\midrule
\multicolumn{3}{c}{\textbf{Severity = 0}} \\
\midrule
PGD           & \cellcolor{gray!20}63.13 & 25.89 \\
TRADES        & \cellcolor{gray!20}55.76 & 27.87 \\
MART          & \cellcolor{gray!20}52.21 & 28.81 \\
DKL           & \cellcolor{gray!20}67.15 & 23.88 \\
\hdashline
\othermethod  & \cellcolor{gray!20}64.03 & 29.60 \\
\methodname   & \cellcolor{gray!20}64.13 & 33.88 \\
\midrule
\multicolumn{3}{c}{\textbf{Severity = 1}} \\
\midrule
PGD           & \cellcolor{gray!20}56.59 & 21.50 \\
TRADES        & \cellcolor{gray!20}53.27 & 21.89 \\
MART          & \cellcolor{gray!20}45.64 & 23.96 \\
DKL           & \cellcolor{gray!20}60.65 & 19.30 \\
\hdashline
\othermethod  & \cellcolor{gray!20}58.21 & 24.51 \\
\methodname   & \cellcolor{gray!20}58.93 & 27.66 \\
\midrule
\multicolumn{3}{c}{\textbf{Severity = 2}} \\
\midrule
PGD           & \cellcolor{gray!20}53.28 & 18.55 \\
TRADES        & \cellcolor{gray!20}47.32 & 18.44 \\
MART          & \cellcolor{gray!20}16.99 & 1.26 \\
DKL           & \cellcolor{gray!20}56.53 & 17.11 \\
\hdashline
\othermethod  & \cellcolor{gray!20}47.69 & 22.50 \\
\methodname   & \cellcolor{gray!20}53.76 & 24.42 \\
\bottomrule
\end{tabular}}
\end{subtable}
}
\end{table*}

The comparison with supervised baselines shows that \methodname{} remains strongly competitive despite not using target labels during adaptation. On full CIFAR-10 (Table~\ref{tab:full_cifar10_supervised}), the proposed method provides the best overall robustness--accuracy trade-off at low and moderate shift, combining high clean accuracy with the strongest robust performance. Under the most severe corruption, PGD performs slightly better, but \methodname{} remains close and still compares favorably with the other supervised objectives.

The same trend is even more evident in the 50/50 CIFAR-10 test-split setting (Table~\ref{tab:split_cifar10_supervised}), which is more challenging and better reflects adaptation with limited target data. In this regime, \methodname{} consistently achieves the strongest robust performance across all severity levels while maintaining competitive clean accuracy. By contrast, supervised methods appear more sensitive in this setting, often showing a sharper degradation in the clean-robustness balance and, in some cases, signs of adversarial overfitting, where robustness on the adaptation data does not translate into equally strong generalization to the test split (e.g., PGD in Table~\ref{tab:split_cifar10_supervised}).
A similar behavior is observed on ImageNet-Val with ResNet-50 (Table~\ref{tab:imagenet_supervised}). Across all corruption levels, \methodname{} achieves the most favorable robustness--accuracy trade-off among the compared methods, including the supervised baselines.

Overall, these results indicate that access to target labels does not necessarily yield a better adaptation strategy under distribution shift. In practice, \methodname{} remains competitive with, and in several cases outperforms, supervised adversarial finetuning baselines, while preserving the advantages of the unsupervised test-time setting.

 \section{Conclusions}
\label{sec:conclusions}
This work studied adversarial robustness adaptation at test time, where a pretrained model must adapt under distribution shift without source data and using only unlabeled target samples. To address this setting, we proposed Teacher-guided Robust Adaptation (\methodname), a label-free framework that uses the predictions of a non-robust teacher as a semantic anchor for both the clean and adversarial objectives during student adaptation. A theoretical analysis showed that self-consistency-based robust adaptation is affected by a moving-target effect that can destabilize optimization, whereas teacher anchoring provides a more stable training signal. Experiments on CIFAR-10 and ImageNet support this view: \methodname consistently improves the robustness--accuracy trade-off over \othermethod, is less sensitive to the choice of $\beta$, and remains effective across architectures, corruption severities, and low-data regimes. Overall, the results indicate that even non-robust pretrained models can provide useful semantic guidance for robust adaptation.

In future work, we plan to further investigate more advanced test-time adaptation techniques for the teacher, with the goal of improving its guidance of the student. This may also open the way to exploiting teacher updates more effectively to further enhance student robustness. Finally, extending \methodname{} to continual and non-stationary adaptation settings, where the target distribution changes over time, represents another promising direction for future research.

 \section*{Acknowledgments}
 This research was partially supported by the Italian National Cybersecurity Agency (Agenzia per la Cybersicurezza Nazionale, ACN),
 and also supported by project SERICS (PE00000014) under the MUR (Ministero dell'Universit\`a e della Ricerca) National Recovery and Resilience Plan funded by the European Union - NextGenerationEU.

 \section*{Declaration on the Use of Generative AI}
 During the preparation of this manuscript, the authors used generative AI tools (ChatGPT 5 and Claude) exclusively to assist with grammar correction and language refinement.

 \bibliographystyle{plain}
 \bibliography{main}

@article{madry2017towards,
  title={Towards deep learning models resistant to adversarial attacks},
  author={Madry, Aleksander and Makelov, Aleksandar and Schmidt, Ludwig and Tsipras, Dimitris and Vladu, Adrian},
  journal={arXiv preprint arXiv:1706.06083},
  year={2017}
}

@inproceedings{zhang2019theoretically,
  title={Theoretically principled trade-off between robustness and accuracy},
  author={Zhang, Hongyang and Yu, Yaodong and Jiao, Jiantao and Xing, Eric and El Ghaoui, Laurent and Jordan, Michael},
  booktitle={International conference on machine learning},
  pages={7472--7482},
  year={2019},
  organization={PMLR}
}

@article{hinton2015distilling,
  title={Distilling the knowledge in a neural network},
  author={Hinton, Geoffrey and Vinyals, Oriol and Dean, Jeff},
  journal={arXiv preprint arXiv:1503.02531},
  year={2015}
}

@inproceedings{furlanello2018born,
  title={Born again neural networks},
  author={Furlanello, Tommaso and Lipton, Zachary and Tschannen, Michael and Itti, Laurent and Anandkumar, Anima},
  booktitle={International conference on machine learning},
  pages={1607--1616},
  year={2018},
  organization={PMLR}
}

@article{pietrosanti2025benchmarking,
  title={Benchmarking the spatial robustness of DNNs via natural and adversarial localized corruptions},
  author={Pietrosanti, Giulia Marchiori and Rossolini, Giulio and Biondi, Alessandro and Buttazzo, Giorgio},
  journal={Pattern Recognition},
  pages={112412},
  year={2025},
  publisher={Elsevier}
}

@article{adapt_SHAMSOLMOALI2021104268,
title = {Advances in domain adaptation for computer vision},
journal = {Image and Vision Computing},
volume = {114},
pages = {104268},
year = {2021},
issn = {0262-8856},
doi = {https://doi.org/10.1016/j.imavis.2021.104268},
url = {https://www.sciencedirect.com/science/article/pii/S0262885621001736},
author = {Pourya Shamsolmoali and Salvador García and Huiyu Zhou and M. Emre Celebi}
}

@article{adapt_NGUYENMEIDINE2021104096,
title = {Knowledge distillation methods for efficient unsupervised adaptation across multiple domains},
journal = {Image and Vision Computing},
volume = {108},
pages = {104096},
year = {2021},
issn = {0262-8856},
author = {Le Thanh Nguyen-Meidine and Atif Belal and Madhu Kiran and Jose Dolz and Louis-Antoine Blais-Morin and Eric Granger},
}

@article{chen2021ltd,
  title={Ltd: Low temperature distillation for robust adversarial training},
  author={Chen, Erh-Chung and Lee, Che-Rung},
  journal={arXiv preprint arXiv:2111.02331},
  year={2021}
}

@article{jeddi2020simple,
  title={A simple fine-tuning is all you need: Towards robust deep learning via adversarial fine-tuning},
  author={Jeddi, Ahmadreza and Shafiee, Mohammad Javad and Wong, Alexander},
  journal={arXiv preprint arXiv:2012.13628},
  year={2020}
}

@inproceedings{zhu2023improving,
  title={Improving generalization of adversarial training via robust critical fine-tuning},
  author={Zhu, Kaijie and Hu, Xixu and Wang, Jindong and Xie, Xing and Yang, Ge},
  booktitle={Proceedings of the IEEE/CVF international conference on computer vision},
  pages={4424--4434},
  year={2023}
}

@article{zagoruyko2016wide,
  title={Wide residual networks},
  author={Zagoruyko, Sergey and Komodakis, Nikos},
  journal={arXiv preprint arXiv:1605.07146},
  year={2016}
}

@article{ge2023boosting,
  title={Boosting adversarial transferability by achieving flat local maxima},
  author={Ge, Zhijin and Liu, Hongying and Xiaosen, Wang and Shang, Fanhua and Liu, Yuanyuan},
  journal={Advances in Neural Information Processing Systems},
  volume={36},
  pages={70141--70161},
  year={2023}
}

@inproceedings{croce2020reliable,
  title={Reliable evaluation of adversarial robustness with an ensemble of diverse parameter-free attacks},
  author={Croce, Francesco and Hein, Matthias},
  booktitle={International conference on machine learning},
  pages={2206--2216},
  year={2020},
  organization={PMLR}
}

@inproceedings{andriushchenko2020square,
  title={Square attack: a query-efficient black-box adversarial attack via random search},
  author={Andriushchenko, Maksym and Croce, Francesco and Flammarion, Nicolas and Hein, Matthias},
  booktitle={European conference on computer vision},
  pages={484--501},
  year={2020},
  organization={Springer}
}

@article{taori2020measuring,
  title={Measuring robustness to natural distribution shifts in image classification},
  author={Taori, Rohan and Dave, Achal and Shankar, Vaishaal and Carlini, Nicholas and Recht, Benjamin and Schmidt, Ludwig},
  journal={Advances in Neural Information Processing Systems},
  volume={33},
  pages={18583--18599},
  year={2020}
}

@article{hendrycks2019benchmarking,
  title={Benchmarking neural network robustness to common corruptions and perturbations},
  author={Hendrycks, Dan and Dietterich, Thomas},
  journal={arXiv preprint arXiv:1903.12261},
  year={2019}
}

@article{amodei2016concrete,
  title={Concrete problems in AI safety},
  author={Amodei, Dario and Olah, Chris and Steinhardt, Jacob and Christiano, Paul and Schulman, John and Man{\'e}, Dan},
  journal={arXiv preprint arXiv:1606.06565},
  year={2016}
}

@article{croce2020robustbench,
  title={Robustbench: a standardized adversarial robustness benchmark},
  author={Croce, Francesco and Andriushchenko, Maksym and Sehwag, Vikash and Debenedetti, Edoardo and Flammarion, Nicolas and Chiang, Mung and Mittal, Prateek and Hein, Matthias},
  journal={arXiv preprint arXiv:2010.09670},
  year={2020}
}

@inproceedings{wang2019improving,
  title={Improving adversarial robustness requires revisiting misclassified examples},
  author={Wang, Yisen and Zou, Difan and Yi, Jinfeng and Bailey, James and Ma, Xingjun and Gu, Quanquan},
  booktitle={International conference on learning representations},
  year={2019}
}

@article{cui2024decoupled,
  title={Decoupled kullback-leibler divergence loss},
  author={Cui, Jiequan and Tian, Zhuotao and Zhong, Zhisheng and Qi, Xiaojuan and Yu, Bei and Zhang, Hanwang},
  journal={Advances in Neural Information Processing Systems},
  volume={37},
  pages={74461--74486},
  year={2024}
}

@article{ngnawe2025robust,
  title={Robust Fine-Tuning from Non-Robust Pretrained Models: Mitigating Suboptimal Transfer With Adversarial Scheduling},
  author={Ngnaw{\'e}, Jonas and Heuillet, Maxime and Sahoo, Sabyasachi and Pequignot, Yann and Ahmad, Ola and Durand, Audrey and Precioso, Fr{\'e}d{\'e}ric and Gagn{\'e}, Christian},
  journal={arXiv preprint arXiv:2509.23325},
  year={2025}
}

@inproceedings{dong2024adversarially,
  title={Adversarially robust distillation by reducing the student-teacher variance gap},
  author={Dong, Junhao and Koniusz, Piotr and Chen, Junxi and Ong, Yew-Soon},
  booktitle={European Conference on Computer Vision},
  pages={92--111},
  year={2024},
  organization={Springer}
}

@article{zhu2021reliable,
  title={Reliable adversarial distillation with unreliable teachers},
  author={Zhu, Jianing and Yao, Jiangchao and Han, Bo and Zhang, Jingfeng and Liu, Tongliang and Niu, Gang and Zhou, Jingren and Xu, Jianliang and Yang, Hongxia},
  journal={arXiv preprint arXiv:2106.04928},
  year={2021}
}

@article{alhamoud2022generalizability,
  title={Generalizability of adversarial robustness under distribution shifts},
  author={Alhamoud, Kumail and Hammoud, Hasan Abed Al Kader and Alfarra, Motasem and Ghanem, Bernard},
  journal={arXiv preprint arXiv:2209.15042},
  year={2022}
}

@inproceedings{sun2020test,
  title={Test-time training with self-supervision for generalization under distribution shifts},
  author={Sun, Yu and Wang, Xiaolong and Liu, Zhuang and Miller, John and Efros, Alexei and Hardt, Moritz},
  booktitle={International conference on machine learning},
  pages={9229--9248},
  year={2020},
  organization={PMLR}
}

@article{wang2020tent,
  title={Tent: Fully test-time adaptation by entropy minimization},
  author={Wang, Dequan and Shelhamer, Evan and Liu, Shaoteng and Olshausen, Bruno and Darrell, Trevor},
  journal={arXiv preprint arXiv:2006.10726},
  year={2020}
}

@article{zhang2022memo,
  title={Memo: Test time robustness via adaptation and augmentation},
  author={Zhang, Marvin and Levine, Sergey and Finn, Chelsea},
  journal={Advances in neural information processing systems},
  volume={35},
  pages={38629--38642},
  year={2022}
}

@inproceedings{wang2022continual,
  title={Continual test-time domain adaptation},
  author={Wang, Qin and Fink, Olga and Van Gool, Luc and Dai, Dengxin},
  booktitle={Proceedings of the IEEE/CVF conference on computer vision and pattern recognition},
  pages={7201--7211},
  year={2022}
}

@article{zhao2023pitfalls,
  title={On pitfalls of test-time adaptation},
  author={Zhao, Hao and Liu, Yuejiang and Alahi, Alexandre and Lin, Tao},
  journal={arXiv preprint arXiv:2306.03536},
  year={2023}
}

@inproceedings{croce2022evaluating,
  title={Evaluating the adversarial robustness of adaptive test-time defenses},
  author={Croce, Francesco and Gowal, Sven and Brunner, Thomas and Shelhamer, Evan and Hein, Matthias and Cemgil, Taylan},
  booktitle={International Conference on Machine Learning},
  pages={4421--4435},
  year={2022},
  organization={PMLR}
}

@inproceedings{park2024medbn,
  title={Medbn: Robust test-time adaptation against malicious test samples},
  author={Park, Hyejin and Hwang, Jeongyeon and Mun, Sunung and Park, Sangdon and Ok, Jungseul},
  booktitle={Proceedings of the IEEE/CVF Conference on Computer Vision and Pattern Recognition},
  pages={5997--6007},
  year={2024}
}

@inproceedings{lo2024adaptive,
  title={Adaptive batch normalization networks for adversarial robustness},
  author={Lo, Shao-Yuan and Patel, Vishal M},
  booktitle={2024 IEEE International Conference on Advanced Video and Signal Based Surveillance (AVSS)},
  pages={1--6},
  year={2024},
  organization={IEEE}
}

@article{wong2020fast,
  title={Fast is better than free: Revisiting adversarial training},
  author={Wong, Eric and Rice, Leslie and Kolter, J Zico},
  journal={arXiv preprint arXiv:2001.03994},
  year={2020}
}

@article{shafahi2019adversarial,
  title={Adversarial training for free!},
  author={Shafahi, Ali and Najibi, Mahyar and Ghiasi, Mohammad Amin and Xu, Zheng and Dickerson, John and Studer, Christoph and Davis, Larry S and Taylor, Gavin and Goldstein, Tom},
  journal={Advances in neural information processing systems},
  volume={32},
  year={2019}
}

@inproceedings{he2016deep,
  title={Deep residual learning for image recognition},
  author={He, Kaiming and Zhang, Xiangyu and Ren, Shaoqing and Sun, Jian},
  booktitle={Proceedings of the IEEE conference on computer vision and pattern recognition},
  pages={770--778},
  year={2016}
}

@article{dosovitskiy2020image,
  title={An image is worth 16x16 words: Transformers for image recognition at scale},
  author={Dosovitskiy, Alexey and Beyer, Lucas and Kolesnikov, Alexander and Weissenborn, Dirk and Zhai, Xiaohua and Unterthiner, Thomas and Dehghani, Mostafa and Minderer, Matthias and Heigold, Georg and Gelly, Sylvain and others},
  journal={arXiv preprint arXiv:2010.11929},
  year={2020}
}

@article{mo2022adversarial,
  title={When adversarial training meets vision transformers: Recipes from training to architecture},
  author={Mo, Yichuan and Wu, Dongxian and Wang, Yifei and Guo, Yiwen and Wang, Yisen},
  journal={Advances in Neural Information Processing Systems},
  volume={35},
  pages={18599--18611},
  year={2022}
}

@article{li2016revisiting,
  title={Revisiting batch normalization for practical domain adaptation},
  author={Li, Yanghao and Wang, Naiyan and Shi, Jianping and Liu, Jiaying and Hou, Xiaodi},
  journal={arXiv preprint arXiv:1603.04779},
  year={2016}
}

@article{schmidt2018adversarially,
  title={Adversarially robust generalization requires more data},
  author={Schmidt, Ludwig and Santurkar, Shibani and Tsipras, Dimitris and Talwar, Kunal and Madry, Aleksander},
  journal={Advances in neural information processing systems},
  volume={31},
  year={2018}
}

@article{goodfellow2014explaining,
  title={Explaining and harnessing adversarial examples},
  author={Goodfellow, Ian J and Shlens, Jonathon and Szegedy, Christian},
  journal={arXiv preprint arXiv:1412.6572},
  year={2014}
}

@article{zeng2024rethinking,
  title={Rethinking precision of pseudo label: Test-time adaptation via complementary learning},
  author={Zeng, Longbin and Han, Jiayi and Du, Liang and Ding, Weiyang},
  journal={Pattern Recognition Letters},
  volume={177},
  pages={96--102},
  year={2024},
  publisher={Elsevier}
}

 \appendix
 \section{Appendix}
\subsection{Robustness under distribution shift.}
\begin{figure}[!htbp]
    \centering
    \begin{subfigure}{0.85\columnwidth}
        \begin{subfigure}{\columnwidth}
        \includegraphics[width=\textwidth]{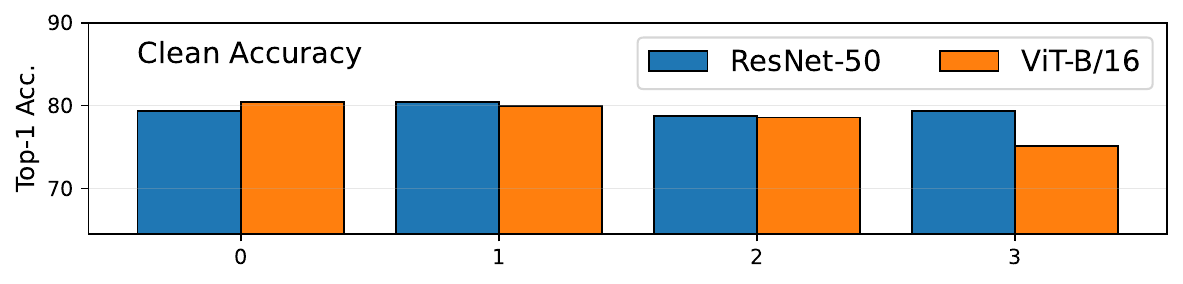}
        \end{subfigure}
        \begin{subfigure}{\columnwidth}
        \includegraphics[width=\textwidth]{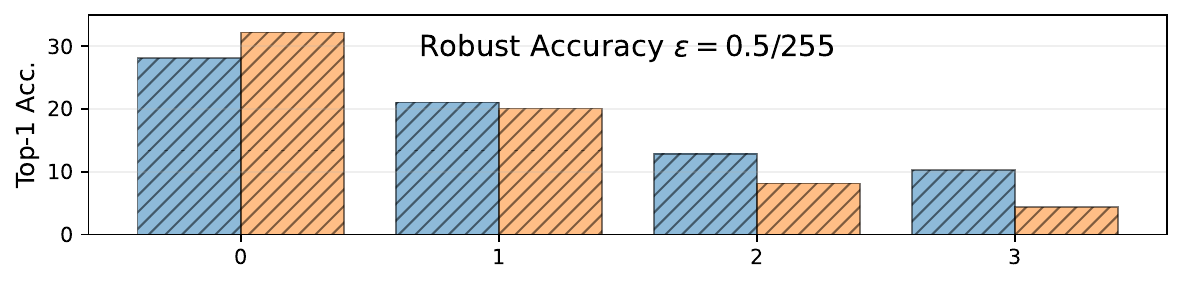}
        \end{subfigure}
        \begin{subfigure}{\columnwidth}
        \includegraphics[width=\textwidth]{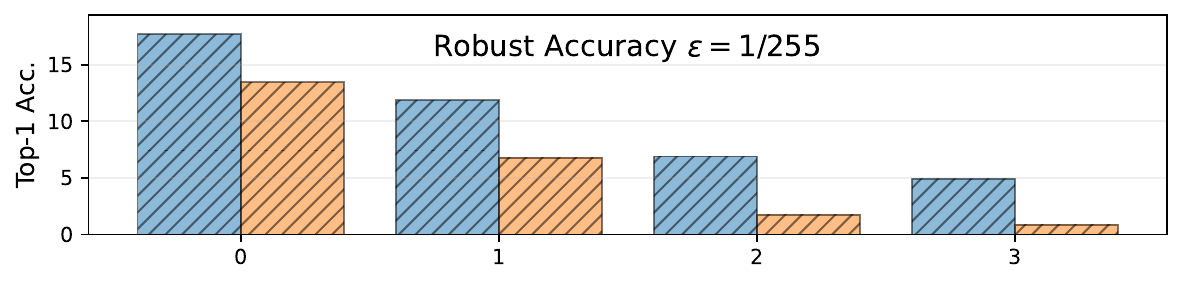}
        \end{subfigure}
        \begin{subfigure}{\columnwidth}
        \includegraphics[width=\textwidth]{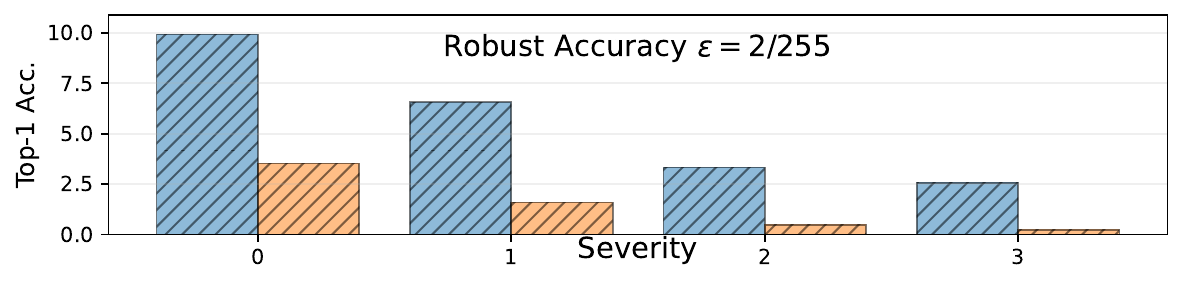}
        \end{subfigure}
    \end{subfigure}
    \caption{\small{
Clean and robust accuracy (Top-1) on ImageNet validation under increasing photometric corruption severity. Robust accuracy is evaluated using PGD-20 with different $\epsilon$. }}
    \label{fig:imagenet_preliminaries}
\end{figure}

As stated in the motivation (Section~\ref{sec:related}), the need to explore robustification techniques in test-time scenarios for well-known pretrained models can also be illustrated through a simple yet effective analysis. As shown in Figure~\ref{fig:imagenet_preliminaries}, even mild photometric transformations applied to common pretrained torchvision models lead to a significant degradation in adversarial robustness across different values of $\epsilon$, while clean performance remains relatively stable. This mismatch highlights that standard generalization does not guarantee robustness under distribution shift.
\subsection{Derivation of Eqs.~\eqref{eq:self_grad_main} and~\eqref{eq:teach_grad_main}.}
\label{sec:derivation_prop}
We derive the gradients of the two robust regularizers introduced in Section~\ref{sec:theory_stability}. 
Let $\hat{x}_\theta$ denote an inner maximizer of the corresponding objective. Under a standard first-order treatment of the inner maximization (e.g., by Danskin's theorem, or equivalently by differentiating the outer objective at a maximizing perturbation), we can write
\[
\nabla_\theta R_{\mathrm{self}}(\theta;x)
=
\nabla_\theta
\mathrm{KL}\!\big(p_\theta(x)\,\|\,p_\theta(\hat{x}_\theta)\big),
\]
and
\[
\nabla_\theta R_{\mathrm{teach}}(\theta;x)
=
\nabla_\theta
\mathrm{KL}\!\big(q(x)\,\|\,p_\theta(\hat{x}_\theta)\big).
\]
For the self-consistency regularizer, define
\(
F(p,p') := \mathrm{KL}(p\,\|\,p').
\)
Then
\(
R_{\mathrm{self}}(\theta;x)
=
F\big(p_\theta(x),\,p_\theta(\hat{x}_\theta)\big).
\)
Applying the chain rule with respect to the two arguments of $F$ yields
\[
\begin{aligned}
\nabla_\theta R_{\mathrm{self}}(\theta;x)
={}&
\big(\partial_\theta p_\theta(x)\big)^\top
\nabla_p F\big(p_\theta(x),p_\theta(\hat{x}_\theta)\big)
\\
&+
\big(\partial_\theta p_\theta(\hat{x}_\theta)\big)^\top
\nabla_{p'} F\big(p_\theta(x),p_\theta(\hat{x}_\theta)\big).
\end{aligned}
\]
Substituting the definition of $F$ gives
\[
\begin{aligned}
\nabla_\theta R_{\mathrm{self}}(\theta;x)
={}&
\big(\partial_\theta p_\theta(x)\big)^\top
\nabla_{p} \mathrm{KL}\!\big(p\,\|\,p_\theta(\hat{x}_\theta)\big)\big|_{p=p_\theta(x)}
\\
&+
\big(\partial_\theta p_\theta(\hat{x}_\theta)\big)^\top
\nabla_{p'} \mathrm{KL}\!\big(p_\theta(x)\,\|\,p'\big)\big|_{p'=p_\theta(\hat{x}_\theta)} .
\end{aligned}
\]
which is Eq.~\eqref{eq:self_grad_main}.

For the teacher-anchored regularizer, we define
\(
G(p') := \mathrm{KL}\!\big(q(x)\,\|\,p'\big).
\)
Since the teacher distribution $q(x)$ is treated as fixed, the regularizer depends on $\theta$ only through the adversarial prediction:
\(
R_{\mathrm{teach}}(\theta;x)
=
G\big(p_\theta(\hat{x}_\theta)\big).
\)
Applying the chain rule gives
\[
\nabla_\theta R_{\mathrm{teach}}(\theta;x)
=
\big(\partial_\theta p_\theta(\hat{x}_\theta)\big)^\top
\nabla_{p'} G\big(p_\theta(\hat{x}_\theta)\big).
\]
Substituting the definition of $G$ yields
\[
\nabla_\theta R_{\mathrm{teach}}(\theta;x)
=
\big(\partial_\theta p_\theta(\hat{x}_\theta)\big)^\top
\nabla_{p'} \mathrm{KL}\!\big(q(x)\,\|\,p'\big)\big|_{p'=p_\theta(\hat{x}_\theta)},
\]
which is Eq.~\eqref{eq:teach_grad_main}.

\subsection{Target-domain corruptions}
\label{app:corruptions}
The target-domain shifts are generated through a fixed compound augmentation composed of Gaussian noise, Gaussian blur, and color jitter, applied in this order. In the experiments, we consider only severity levels $s \in \{1,2\}$.

More precisely, given an input image $x$, the corrupted sample is obtained as
\(
\tilde{x} =
\mathcal{T}^{(s)}_{\text{jitter}}
\circ
\mathcal{T}^{(s)}_{\text{blur}}
\circ
\mathcal{T}^{(s)}_{\text{noise}}(x).
\)
For Gaussian noise, we use additive perturbations with standard deviation $\sigma_s$, namely $\sigma_1=0.03$ and $\sigma_2=0.06$. Blur is applied with kernel size $k_1=3$ and $k_2=5$. Color jitter modifies brightness, contrast, and saturation with the same strength $a_s$, with $a_1=0.1$ and $a_2=0.2$, while hue is fixed to $0$.

\begin{figure}[!htbp]
    \centering
    \captionsetup{font=small}
    \captionsetup[subfigure]{font=small, justification=centering}
    \begin{subfigure}{0.9\linewidth}
        \centering
        \includegraphics[width=\linewidth]{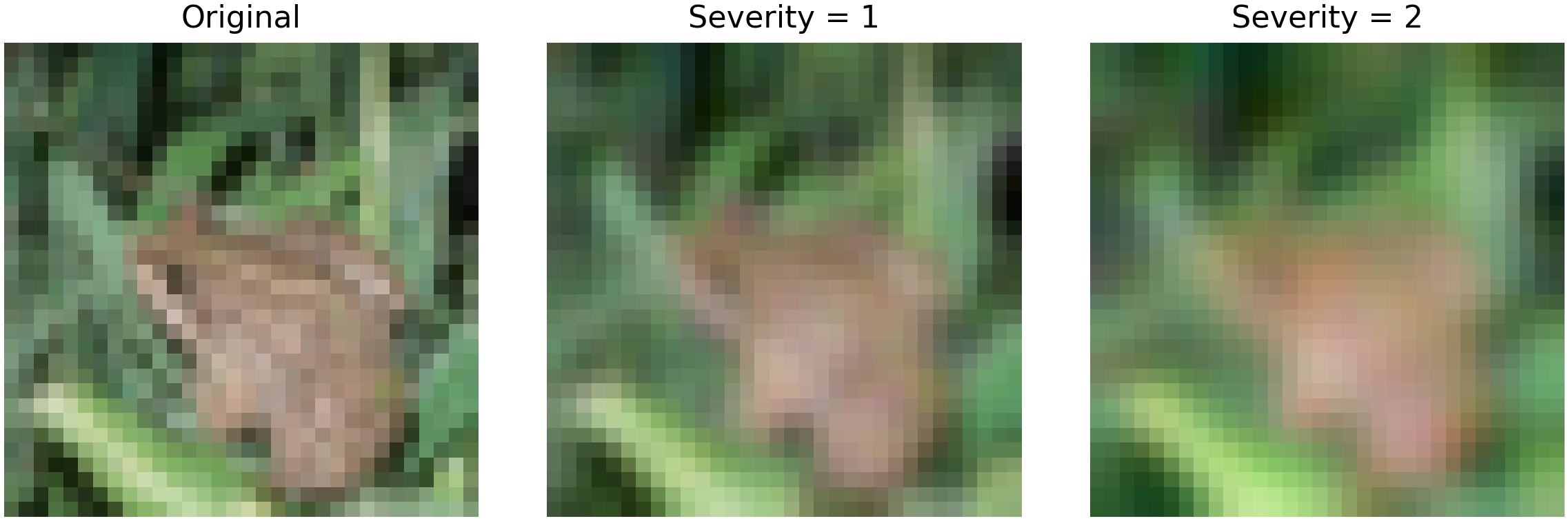}
        \caption{CIFAR-10}
    \end{subfigure}
    \begin{subfigure}{0.9\linewidth}
        \centering
        \includegraphics[width=\linewidth]{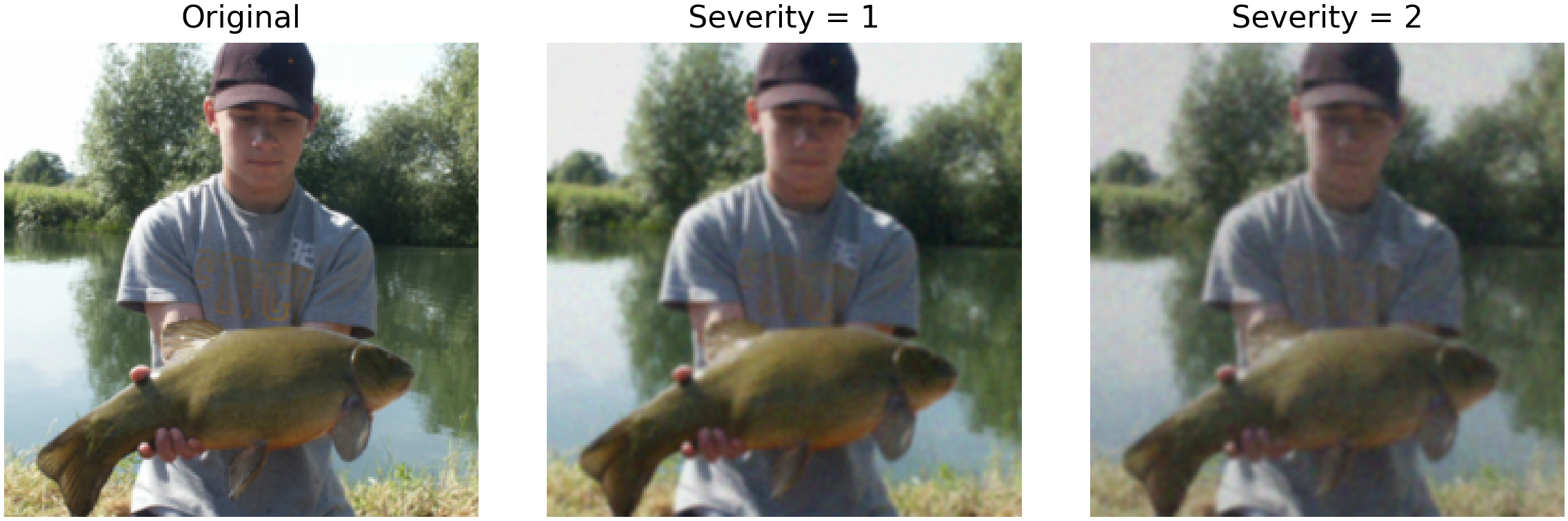}
        \caption{ImageNet}
    \end{subfigure}
    \caption{Examples of the target-domain corruptions for CIFAR-10 (top) and ImageNet (bottom).}
    \label{fig:corruption-examples}
\end{figure}

A representative corrupted example for CIFAR-10 and one for ImageNet are shown in Figure~\ref{fig:corruption-examples}. Table~\ref{tab:corruption-details} summarizes the exact corruption settings.
\begin{table}[t]
\centering
\caption{Compound corruption settings used to define the target-domain shifts.}
\label{tab:corruption-details}
\begin{tabular}{c|c|c|c}
\toprule
\textbf{Severity} & \textbf{Noise }$\sigma$ & \textbf{Blur kernel} & \textbf{Jitter }$(b,c,s,h)$ \\
\midrule
1 & 0.03 & 3 & (0.1, 0.1, 0.1, 0) \\
2 & 0.06 & 5 & (0.2, 0.2, 0.2, 0) \\
\bottomrule
\end{tabular}
\end{table}

\subsection{Source pretrain models setup}
\label{sec:source_pretraining}
Pre-training is performed using SGD with momentum $0.9$ and weight decay $5\times 10^{-4}$ for $200$ epochs. The initial learning rate is set to $0.1$ and decayed by a factor of $0.2$ at epochs $60$, $120$, and $160$. A dropout rate of $0.3$ is used during pretraining.
For ImageNet, we consider two pretrained torchvision architectures: ResNet-50 \cite{he2016deep} initialized with \texttt{ResNet50\_Weights} \texttt{.IMAGENET1K\_V2}, and ViT-B/16 \cite{dosovitskiy2020image} initialized with \texttt{ViT\_B\_16\_Weights} \texttt{.IMAGENET1K\_V1}. These pretrained models are used to initialize the teacher/student setup before target-domain adaptation on the split validation data.

\subsection{Analysis of limited data on ImageNet-Val}
\begin{table}[h]
\centering
\caption{\small{Analysis of the effect of test-time adaptation dataset size on ResNet50 and the ImageNet validation-set split, where the first value denotes the finetuning set size and the second the test set size, with no data corruption in either the finetuning or test data.}}
\resizebox{0.45\textwidth}{!}{%
\begin{tabular}{l c || c c || c c}
\toprule
\multirow{2}{*}{Split} & \multirow{2}{*}{$\varepsilon$} & \multicolumn{2}{c||}{Clean} & \multicolumn{2}{c}{Robust} \\
\cline{3-6}
 &  & Top-1 & Top-5 & Top-1 & Top-5 \\
\midrule
50/10 & \multirow{4}{*}{$2/255$} & 65.42 & 85.90 & 35.34 & 64.40 \\
70/10 &  & 66.56 & 86.64 & 37.30 & 66.32 \\
80/10 &  & 66.20 & 86.74 & 38.22 & 66.68 \\
90/10 &  & 68.02 & 87.66 & 34.56 & 62.78 \\
\midrule
50/10 & \multirow{4}{*}{$4/255$} & 57.86 & 79.84 & 16.84 & 39.82 \\
70/10 &  & 58.80 & 80.32 & 20.06 & 44.26 \\
80/10 &  & 58.64 & 80.90 & 20.70 & 44.66 \\
90/10 &  & 58.36 & 80.38 & 20.90 & 44.66 \\
\bottomrule
\end{tabular}
}%
\label{tab:ablation_split_imagenet}
\end{table}
We further analyze the effect of target sample availability on ImageNet by varying the portion of the validation set used for test-time finetuning. For this experiment, we focus on the proposed \methodname{} with ResNet-50, and no data corruption is applied, so as to independently study the impact of data availability on the proposed technique during test-time finetuning. The results in Table~\ref{tab:ablation_split_imagenet} confirm that increasing the amount of target data is generally beneficial, especially for robustness.
However, under the smaller perturbation budget ($\varepsilon = 2/255$), we observe that the improvement is not always strictly monotonic: larger splits tend to improve robustness up to an intermediate regime, while the largest split mainly benefits clean accuracy. This trend may depend on the specific quality and composition of the available data, and may therefore not generalize to settings with additional target samples. In contrast, under the stronger perturbation budget, the trend is more regular, with robust accuracy steadily improving as more target samples are used, while clean accuracy remains relatively stable. Overall, these results are consistent with the CIFAR-10 analysis, further confirming that target-data availability plays a central role in robust adaptation.

\end{document}